\documentclass{article} 
\usepackage{collas2026_conference,times}
\usepackage{easyReview}

\usepackage{amsmath,amsfonts,bm}

\def\eqref#1{equation~\ref{#1}}

\def\1{\bm{1}}

\def\vtheta{{\bm{\theta}}}

\def\vx{{\bm{x}}}

\DeclareMathAlphabet{\mathsfit}{\encodingdefault}{\sfdefault}{m}{sl}
\SetMathAlphabet{\mathsfit}{bold}{\encodingdefault}{\sfdefault}{bx}{n}

\usepackage{hyperref}
\hypersetup{
    colorlinks=true,
    linkcolor=red,
    filecolor=magenta,
    urlcolor=blue,
    citecolor=purple,
    pdftitle={Overleaf Example},
    pdfpagemode=FullScreen,
    }
\usepackage{array}
\usepackage{siunitx}
    
\usepackage{booktabs}
\usepackage{amsmath}
\usepackage{amssymb}
\usepackage{amsthm}
\usepackage{subcaption}
\usepackage{xcolor}
\usepackage{algorithm}
\usepackage{algorithmic}
\usepackage{multirow}
\usepackage{cleveref}
\usepackage{pifont}
\usepackage{wrapfig}
\usepackage{silence}
\usepackage{xspace}
\usepackage{caption}
\usepackage{setspace}
\usepackage{tcolorbox}

\usepackage{enumitem}
\usepackage{fmtcount}

\definecolor{myred}{RGB}{187, 1, 54}
\definecolor{mygreen}{RGB}{61, 120, 56}
\definecolor{othergreen}{rgb}{0.0, 0.5, 0.0}
\definecolor{fgold}{RGB}{212, 175, 55}
\definecolor{ssilver}{RGB}{192, 192, 192}
\definecolor{tbronze}{RGB}{159, 122, 52}

\newcommand{\cmark}{\textcolor{mygreen}{\ding{51}}}%
\newcommand{\xmark}{\textcolor{myred}{\ding{55}}}%

\newcommand{\licon}{\includegraphics[scale=0.7]{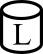}\xspace}
\newcommand{\uicon}{\includegraphics[scale=0.7]{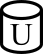}\xspace}
\newcommand{\method}{I\&P\xspace}
\newcommand{\methodlong}{\emph{Improve \& Prune}\xspace(I\&P)\xspace}
\newcommand{\minisection}[1]{\noindent {\bf #1}\ }

\title{One Loop, Two Gains: Can Active Learning\\win the Lottery for Free?}

\author{Benedikt Tscheschner \\
University of Technology Graz \\
Know-Center Research GmbH \\
Austria \\
\texttt{btscheschner@know-center.at} \\
\And 
Eduardo Veas  \\
University of Technology Graz \\
Know-Center Research GmbH \\
Austria \\
\texttt{eveas@know-center.at} \\
\And 
Marc Masana \\
Institute of Visual Computing \\
University of Technology Graz \\
Austria \\
\texttt{mmasana@tugraz.at}
}

\collasfinalcopy 

\begin{document}
\maketitle

\begin{abstract}
    The lottery ticket hypothesis posits the existence of \emph{winning tickets}: sparse subnetworks that, when trained in isolation from their original initialization, match the accuracy of the full dense network. The predominant method for discovering such tickets, iterative magnitude pruning, alternates pruning with full retraining from scratch until convergence over many cycles. Similarly, deep active learning also retrains a model from scratch after each acquisition round as new labels become available. Despite this shared reliance on iterative retraining with a substantial computational overhead, the two paradigms have been studied separately. We observe that the iterative training loop inherent to pool-based active learning already provides the exact computational structure that iterative magnitude pruning exploits, and propose \methodlong, a method that integrates magnitude pruning into each active learning retraining cycle at practically no additional cost. This raises a key empirical question: can iterative magnitude pruning produce winning tickets under the non-stationary data regime of active learning? We investigate this question across multiple acquisition functions, architecture families, and image classification datasets, including an active fine-tuning scenario. Our results demonstrate that \method yields sparse, deployable models at each active learning iteration. Those match the accuracy of their dense counterparts at sparsities up to 95\%, effectively obtaining winning tickets as a byproduct of the active learning pipeline. These per-iteration sparse models can address two computational bottlenecks --- per-round model retraining and acquisition scoring over the unlabeled pool --- that currently prevent the practical adoption of DAL on large architectures and large unlabeled pools.
\end{abstract}

\section{Introduction}\label{sec:intro}
Training deep neural networks demands both large labeled datasets and large models, yet practitioners in specialized domains such as medical imaging or industrial analysis often face a scarcity of expert annotations~\citep{settles2009active, munro2021human}. In these domains, models are often deployed on portable diagnostic devices or on-premise industrial hardware~\citep{han2015learning} under tight memory, energy, and latency constraints, and are queried at scale once in production, making inference-time compute a first-class cost alongside annotation effort. Deep active learning~(DAL)~\citep{li2024survey} and iterative magnitude pruning~(IMP)~\citep{frankle2019stabilizing} tackle these two costs independently. DAL minimizes annotation cost by iteratively selecting the most informative samples to be labeled from a large unlabeled pool. IMP discovers highly sparse subnetworks, so-called \emph{winning tickets}~\citep{frankle2018the}, that match dense-model accuracy at a fraction of the parameters through repeated cycles of training and pruning. Although the two paradigms target different resources --- annotations and model size respectively --- they share a common computational backbone: both retrain the model from scratch over multiple rounds. This structural overlap suggests a natural, yet unexplored, opportunity to combine them --- one that becomes more important as DAL scales to modern workloads, where per-round retraining grows costly on large backbones~\citep{das2023accelerating} and acquisition functions must repeatedly score increasingly large unlabeled pools~\citep{tsvigun-etal-2022-plasm, citovsky2021batch}.

We observe that the retraining loop present in pool-based active learning provides exactly the iterative, train-then-prune structure that IMP requires. However, trivially applying the IMP at each DAL iteration would scale the training costs quadratically. Therefore, we want to exploit that DAL already retrains the model from scratch to avoid the optimization bias introduced by warm-starting on a previously smaller dataset~\citep{ash2020warm}). We argue that a single pruning step appended to each round can progressively discover winning tickets at no additional training cost. These winning tickets uncovered along the way are not merely a deployment side effect --- they directly alleviate a key scalability bottleneck inside the DAL loop itself~\citep{citovsky2021batch}. 

Acquisition functions must score the entire unlabeled pool, often repeatedly in Bayesian approaches~\citep{gal2017deep}, which makes inference costs scale directly with pool size. Sparse models reduce both the number of per-pass FLOPs and the effective embedding dimension used by distance-based acquisition functions, making the DAL loop scale more favorably to large pools.

We formalize this integration of both paradigms as \methodlong, a drop-in extension to any pool-based DAL pipeline that yields sparse models at each DAL iteration. Unlike the original IMP, our method operates on a non-stationary labeled set. To accommodate for this, we relax the strict monotonic shrinking of IMP and let the pruning mask dynamically expand and contract at each iteration. Parameters deemed unimportant on previous iterations can be reactivated while a subsequent pruning step reinstates sparsity. We empirically show that winning tickets nonetheless emerge under this regime across six acquisition functions, four image classification datasets (including an active fine-tuning scenario), and two architecture backbones under both supervised and self-supervised pretraining.

\section{Background}

\subsection{Minimal labeling with deep active learning}\label{sec:pre_al}
DAL aims at efficient deep neural network training through iterative batch-wise sample selection with minimal labeling effort~\citep{munro2021human, li2024survey}. Pool-based active learning, where all data samples are available and fixed from the start, is the dominant setting and the focus of this work; stream-based alternatives, where new unlabeled data arrives over time, are left for future investigation.

For a pool-based scenario, let \mbox{$D^{(0)}_{U}\!\sim\!P$} denote an unlabeled pool drawn from a data distribution $P$, accompanied by an initial set of oracle-labeled samples \mbox{$D^{(0)}_{L}\!\sim\!P$} with \mbox{$D^{(0)}_{U}\cap D^{(0)}_{L}=\emptyset$}. The goal is to learn a model $f(\vx;\vtheta)$ that generalizes over $P$ while querying as few oracle labels from $D^{(0)}_{U}$ as possible. The model is initially trained on $D^{(0)}_{L}$, starting from either a random initialization or a predefined checkpoint $\vtheta_{\text{init}}$. At each iteration $(j)\!>\!0$, an acquisition function $\mathcal{S}$ scores $D_U^{(j)}$ using the embedding representations and pseudo labels of $f(\vx ; \vtheta^{(j)})$ to select a batch of \emph{informative} samples $D_A^{(j)} \subset D_U^{(j)}$. Then, annotations from an oracle are requested for $D_A^{(j)}$ and the partitions are updated as $D^{(j+1)}_{L} = D^{(j)}_{L} \cup D^{(j)}_{A}$ and $D^{(j+1)}_{U} = D^{(j)}_{U} \setminus D^{(j)}_{A}$. A new model is retrained from the initial weights $\theta_{\text{init}}$ on $D_L^{(j+1)}$ to convergence, keeping hyperparameters constant across iterations. In practice, DAL iterates until a predetermined annotation or computation budget is exhausted or a target performance level is reached~\citep{li2024survey}. Furthermore, to avoid a cold start~\citep{schein2002methods, yehuda2022active}, $D^{(0)}_{L}$ is commonly seeded by random draws or by clustering pretrained embeddings so that the initial batch already covers $P$ broadly~\citep{li2024survey}.

The design of the acquisition function, which ranks unlabeled samples for labeling, has been a central research question in deep active learning. The most direct extension of active learning to deep neural networks estimates predictive uncertainty for individual samples and selects those for which the model is least confident~\citep{settles2009active, gal2017deep, kirsch2019batchbald}. While computationally efficient on a per-sample basis, uncertainty-based selection is prone to redundancy within a single acquisition batch and exhibits sensitivity to outliers and noisy labels~\citep{li2024survey}. 

Rather than scoring individual samples, coreset methods aim to select batches that best represent the underlying data distribution $P$, drawing on ideas from clustering and covering theory~\citep{sener2017active, yehuda2022active}. A distance or covering metric is computed over embedding representations of the neural network. Because exact batch selection is often computationally intractable, greedy approximations or random subsampling are standard. 

A third family aims to combine uncertainty and diversity signals in a single function. Some methods estimate the influence of candidate annotations on model parameters, e.g.\ through gradient embeddings or Fisher information, and enforce batch diversity over these representations~\citep{ash2019deep, ash2021gone}. Others pair conventional uncertainty measures with embedding-based clustering to balance informativeness and coverage~\citep{citovsky2021batch, maekawa2022low}. 

No single acquisition function dominates across all settings: different label budgets and dataset sizes favor different strategies~\citep{yehuda2022active}, and recent research has shifted towards selecting the most suitable acquisition function adaptively throughout the DAL loop~\citep{zhang2023algorithm, wang2025autoal, huseljic2026boss}.
To ensure generality, we benchmark against six representative acquisition functions covering all three families since \method imposes no assumptions on $\mathcal{S}$.

\newpage
\subsection{Promoting sparsity with the Lottery Ticket Hypothesis}\label{sec:pre_lth}

\citet{frankle2018the} introduced the Lottery Ticket Hypothesis (LTH):
\begin{tcolorbox}[boxsep=1pt,left=3pt,right=3pt,top=2pt,bottom=2pt]\label{thm:lth}
    \textbf{Lottery Ticket Hypothesis:} A randomly-initialized, dense neural network contains a subnetwork that, when trained in isolation from its original initialization, can match the test accuracy of the original network after training for at most the same number of iterations.
\end{tcolorbox}

The primary method for identifying such \emph{winning tickets} is Iterative Magnitude Pruning (IMP): the model is trained from its original initialization, a fixed fraction of the lowest-magnitude weights are pruned (typically 20\% per round), and the process repeats. The aim is to empirically find the maximum sparsity at which performance begins to degrade within some predefined margin $\epsilon$. 

Let $f(\vx ; \vtheta_{\text{init}})$ be the initial model that is trained on dataset \mbox{$D_{B}\!\sim\!P$} until convergence for $j$ training iterations. A pruning mask for the model is defined as \mbox{$M\!\in\!\{0, 1\}^{|\vtheta|}$} with an initialization \mbox{$M^{(0)}\!=\!1$}, and with a pruning rate $p_{\text{c}}$. A pruning function $\mathcal{C}$ (e.g.\ weight magnitude) ranks the converged parameter set $\theta^{(j)}$. The mask for the next round $M^{(j+1)}$ is obtained by pruning the $p_{\text{c}}$ lowest-ranked weights. At each iteration $(j) \geq 0$, the model \mbox{$f(\vx ; \vtheta_{\text{init}} \, \odot \, M^{(j)})$} trains from scratch with the pruning mask $M^{(j)}$ as a constraint.
After $j$ iterations, IMP yields a family of candidate tickets $(\vtheta_{\text{init}}, M^{(i)})_{i \leq j}$ with geometrically increasing sparsities $1 - (1 - p_{\text{c}})^{i}$; the winning ticket is the sparsest candidate still matching the dense reference within a performance margin $\epsilon$.

\citet{frankle2019stabilizing, pmlr-v119-frankle20a} subsequently showed that pruning alone is insufficient for reliable ticket discovery in large-scale networks. Rewinding weights to an early training epoch rather than to initialization yields more stable winning tickets, especially in deeper architectures. This finding suggests that early optimization dynamics encode critical inductive biases that determine subnetwork viability. The LTH is therefore not merely a pruning heuristic but evidence that plasticity~\citep{frankle2020early, pinson2026itslotteryitsrace} for a given task is unevenly distributed within model parameter space, a perspective we revisit in \Cref{sec:limitations}.

Despite its effectiveness, IMP remains computationally expensive due to the repeated training cycles it requires, and various strategies for reducing this burden have been proposed~\citep{liu2024survey}, including early stopping for ticket discovery~\citep{you2020drawing} and training on a specialized coreset~\citep{zhang2021efficient}. This overhead has largely confined IMP to the role of an analysis tool rather than a practical deployment recipe. However, we observe that this cost is only expensive in isolation: whenever a training pipeline \emph{already} performs repeated retraining (such as our proposed \method), IMP can be layered on top.

\subsection{Pruning for computational savings}
The magnitude signal IMP exploits is only available after convergence, which explains both its reliability and its cost. Pruning methods of neural networks more broadly differ in the training stage at which they commit to an importance estimate~\citep{cheng2024survey}, trading signal quality for computational savings. Some methods determine which weights can be removed from the randomly initialized network before training begins: SNIP~\citep{lee2019snip} measures connection sensitivity with respect to the loss, GraSP~\citep{wang2020picking} preserves gradient flow, and PX~\citep{Iurada_2024_CVPR} aligns the sparse network's training dynamics with those of the dense model via NTK spectral properties. All three require a batch of training data for their estimates, whereas SynFlow~\citep{tanaka2020pruning} operates entirely data-free by conserving synaptic flow to avoid layer collapse. Despite these differences, all methods share a fundamental limitation: parameter importance at initialization is a poor proxy for importance after training, causing consistent underperformance relative to iterative pruning at high sparsity~\citep{Iurada_2024_CVPR, cheng2024survey}.

Rather than committing to a fixed topology upfront, dynamic sparse training methods jointly learn weights and sparse structure in a single training run. RigL~\citep{evci2020rigging} and DSR~\citep{dsr_pruning} maintain a fixed sparsity level throughout training, periodically updating the topology by pruning low-magnitude weights and regrowing connections identified via instantaneous dense gradient computations. While this reduces memory and FLOPs for forward passes, the periodic dense gradient steps and additional hyperparameters introduce optimization complexity~\citep{cheng2024survey}. Finally, there are also post-hoc pruning methods. Most prominently, magnitude pruning~\citep{han2015learning} first trains a dense model to convergence and then removes parameters. The pruning signal is strong since it reflects learned importance, and the post-hoc nature of the approach makes it flexible: it can be applied to any trained model without modifying the training procedure. However, the full dense training cost has already been paid and efficiency gains are realized only at deployment. Iterative pruning generally outperforms single-shot pruning at high sparsity~\citep{wojnar2024how}. Importance scores shift as training progresses, and iterative schemes account for this drift by reassessing importance after each round. IMP is one realization of this principle, operating post-hoc by pruning after convergence and rewinding.

\begin{figure}[t]
  \centering
  \includegraphics[width=0.75\linewidth]{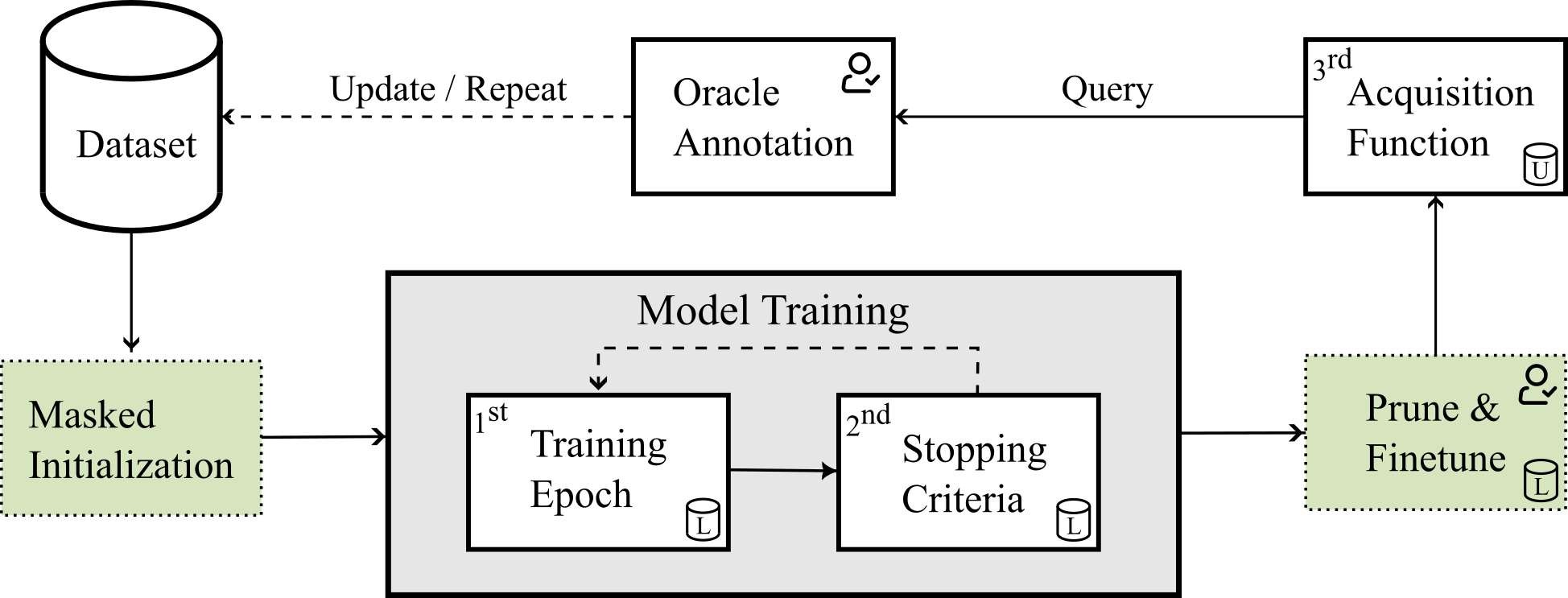}
  \caption{Depiction of a standard DAL loop (white boxes), where the dominant computational costs are model training and acquisition scoring over the unlabeled pool. \method extends DAL by incorporating masked initialization before training and pruning with finetuning after convergence (green boxes). The \licon and \uicon icons indicate whether an operation acts on the labeled or unlabeled set.
  }
  \label{fig:pipeline}
\end{figure}

\section{One loop, two gains}\label{sec:method}

The iterative training by DAL and IMP incurs significant computational costs due to models being repeatedly trained from scratch until convergence. We reduce these scaling costs in DAL by leveraging the increased efficiency of sparse models induced by the LTH. We first analyze the computational scaling, before describing our proposed method.

\subsection{Scaling of Computation in DAL}
The computational cost of a deep active learning pipeline decomposes into
three primary components:

\begin{enumerate}[label=\ordinalnum{\value{enumi}}]
    \item \textit{Model size and sparsity.}
    The FLOPs per forward and backward pass are governed by both the number of
    parameters and the types of operations in the architecture. Because every
    training epoch and every pool inference call pays this cost, reducing model
    complexity through pruning can compound savings across both the training and
    acquisition phases of the DAL loop.

    \item \textit{Optimization strategy and training duration.}
    Per-iteration costs are majorly influenced by how the model is optimized and how long it trains until convergence, which relies on the choice of
    optimizer and momentum, learning rate and its schedule, the number of epochs,
    and early stopping criteria. To avoid unnecessary epochs, we adopt the well established accuracy-based early stopping~\citep{ash2019deep} in our experiments.

    \item \textit{Pool inference and acquisition function.}
    Acquisition functions often require one forward pass over the full unlabeled pool $D_{U}$. Bayesian approaches such as BALD~\citep{gal2017deep} multiply this cost by the number of stochastic passes. Distributional and hybrid methods additionally compute pairwise distances in the embedding space, introducing costs that scale quadratically with the representation dimension. Scaling to large pools remains a practical bottleneck~\citep{citovsky2021batch}, forcing several acquisition families~\citep{saal,bae2025uncertainty} to resort to random subsampling of the unlabeled pool to remain tractable.
\end{enumerate}

This work targets the first and third component through network sparsity; the second is highly task-, dataset-, and architecture-dependent, limiting general prescriptions. Together, these two components form the dominant practical barrier to adopting pool-based DAL on large-scale workloads: per-round retraining from scratch becomes prohibitive as backbone complexity and training data grows~\citep{das2023accelerating}, and acquisition over large unlabeled pools compounds the cost at every iteration~\citep{tsvigun-etal-2022-plasm}. 

\subsection{Improve and Prune}\label{sec:method_ip}
Pool-based active learning already retrains the model from scratch at every acquisition round to avoid warm-start bias~\citep{ash2020warm}. IMP requires an analogous iterative training loop: train to convergence, prune, repeat. \method exploits this structural correspondence by attaching a single pruning step after each DAL training phase (right green box in \Cref{fig:pipeline}), thereby discovering winning lottery tickets at minimal additional cost. At each iteration $(j)$, the model is pruned and the dataset pool grows. This introduces the risk of pruning connections that are required in future iterations, which we address with controlled expansion before training (left green box in \Cref{fig:pipeline}).

Our goal is to iteratively learn a model $f(\vx;\vtheta)$ on $D^{(j)}_{L}\!\sim\!P$ that generalizes over data distribution $P$ while querying as few oracle labels as possible, matching the general setup of DAL described in \Cref{sec:pre_al}. From the IMP perspective, we have \mbox{$D_{B}\!=\!D^{(j)}_{L}$} be the training dataset, which in the DAL setup is non-stationary w.r.t the iterations, and we initialize a mask $M^{(0)}=\textbf{1}$ and a pruning rate $p_{\text{c}}$ as described in \Cref{sec:pre_lth}. Further, we also define a percentage of the total parameters $p_{\text{r}}$ as the reactivation rate, and a minimum percentage of active parameters $(1 - s)$. Next, we first provide a compact definition of our proposed method, and then we expand individually on each of its components.

At iteration $(j)\!>\!0$, we start by expanding the mask in order to restore model capacity. We define a function $\mathcal{R}(M^{(j)}, p_{\text{r}})$ that restores capacity proportionally across layers by setting random mask entries to \num{1}. Then, the model $f(\vx, \vtheta_{\text{init}} \odot \mathcal{R}(M^{(j)}; p_{\text{r}}))$ is trained on $D^{(j)}_{L}$ until convergence. After the training, we apply a pruning function $\mathcal{C}(M^{(j)}; p_{\text{c}})$, which ranks the learned parameters $\vtheta^{(j)}$ and updates the mask by setting $p_{\text{c}}$ mask entries to zero according to the ranking from $\mathcal{C}$, and subject to retaining at least a fraction of $(1 - s)$ active parameters. The model parameters are pruned with the new mask, and finetuned for $\varphi$ epochs or until convergence to compensate for any changes in the old mask pruned parameters. Finally, the acquisition function $\mathcal{S}$ scores the $D^{(j)}_{U}$ based on the sparse model to select informative samples, obtain labels from the oracle and concludes an iteration cycle by updating $D^{(j+1)}_{L}$ and $D^{(j+1)}_{U}$. A pseudo-code version of the described algorithm is available in \Cref{app:Algorithm}.

\minisection{Masked initialization.}
The expansion function $\mathcal{R}$ follows a layer-wise budget, distributing restored capacity proportionally across layers. The relaxed pruning mask remains unstructured, inspired by evolutionary capacity principles~\citep{dsr_pruning}. The rationale is that the labeled set grows by the newly acquired batch $D^{(j)}_{A}$, so parameters pruned under the previous data distribution may carry renewed importance for fitting the expanded dataset; relevant ones will survive the current pruning cycle while unimportant ones will be pruned away. This flexibility is essential: a single mask discovered once and frozen for the remaining iterations gradually drifts out of alignment with the growing labeled set, as we show empirically in \Cref{fig:transfer_mask}.

\minisection{Model Training.}
Our main training stage is equivalent to a standard DAL pipeline: the model is optimized on $D^{(j)}_{L}$ until convergence. The only difference is that training benefits from the sparsity pattern imposed by $M^{(j)}$, which is held \emph{fixed} throughout the training. Therefore, this sparsity is amenable to static sparse-kernel optimization and hardware-level acceleration (e.g.\ 2:4 structured or tiling~\citep{zhao2024beyond}). Dynamic sparse training methods such as RigL~\citep{evci2020rigging} forfeit this opportunity since their masks evolve during training and their regrowth step requires dense gradient computations, preventing the training loop from fully exploiting the nominal FLOP savings.

\minisection{Prune and Finetune.}
Once training has converged, $\mathcal{C}$ is applied to the converged parameters, removing the lowest-magnitude weights at rate $p_{\text{c}}$ up to the target sparsity $s$. Pruning induces an abrupt discontinuity in the function realized by the network leaving the remaining parameters briefly out of alignment with the task. The finetuning-phase~\citep{mishra2021accelerating} continues training on $D^{(j)}_{L}$ under identical conditions, but with an updated mask $M^{(j+1)}$, so that remaining parameters can realign to the unchanged objective. This phase incurs minimal computational overhead since the majority of optimization is performed during model training on the larger mask and the finetuning is restricted to a maximum budget $\varphi$. Therefore, finetuning typically converges within a small fraction of the original training epochs.  Crucially, the finetuning being inexpensive enables efficient exploration of the minimum active parameters $s$. Multiple sparsity levels can be evaluated by branching from the same model training checkpoint and running only the short finetuning phase for each candidate, making the selection of $s$ practical even in resource-constrained settings. This property lends itself naturally to a human-in-the-loop workflow~\citep{munro2021human}: after the first stage of any DAL iteration, a practitioner can branch at multiple sparsity levels, inspect the resulting efficiency trade-off on a test set, and select the operating point that best fits the deployment constraints before committing to the next acquisition round.

\minisection{Acquisition.} The acquisition step, operating over the large unlabeled pool $|D^{(j)}_U| \gg |D^{(j)}_L|$, benefits from the refined sparsity of the finetuned model, in terms of generalization to distribution shifts, better calibrated predictive uncertainty and computational efficiency~\citep{chen2022can}. Any distance-based $\mathcal{S}$ can gain a quadratic speedup from the reduced latent representation dimension without requiring specialized sparse operations (see Appendix~\ref{app:flops_scaling} for scaling details).

\bigskip
\method is agnostic to both the reactivation and pruning criteria. This work adopts unstructured magnitude pruning for $\mathcal{C}$ following the original LTH formulation~\citep{frankle2018the}, and random layer-wise reactivation for $\mathcal{R}$~\citep{dsr_pruning}.
These can be formally defined as,
\begin{equation}\label{eq:cr_instantiation}
    \mathcal{C}\!\left(M^{(j)}, \vtheta^{(j)}\right) = M^{(j)} \odot 1 \! \left[\,|\vtheta^{(j)}| > \tau^{(j)}\,\right], \qquad
    \mathcal{R}\!\left(M^{(j)}\right) = M^{(j)} \,\lor\, \xi^{(j)}\,,
\end{equation}
where $\tau^{(j)}$ is the threshold that removes the lowest $p_{\text{c}}$ fraction of active weights and $\xi^{(j)}$ is a layer-stratified random mask that activates a $p_{\text{r}}$ fraction of pruned weights. Conceptually, DAL and IMP could be run as two nested loops, paying the iterative training cost of IMP on top of the cost already incurred by DAL. \method fuses these into one loop with two gains: progressively sparser models and label-efficient acquisition, without extra winning ticket search cost.

\begin{table}[t]
    \centering
    \caption{\textit{Acquisition does not affect pruning.} Final classification accuracy $A_{9}$ [\%] for a \mbox{ResNet-18} pretrained with self-supervised SimCLR on CIFAR-100 and Imagewoof under a low (LB) and high (HB) annotation budget. Rewinding points are taken at half of the warmup period. Results report the mean $\pm$ std over \num{10} seeds. $p$-value of one-sided Wilcoxon test for non-inferiority against Dense on AULC; \cmark\ if $p<0.05$ (margin $\epsilon=0.005$), \xmark\ otherwise.
    }
    \label{tab:final_acc_results}
    \resizebox{\linewidth}{!}{%
    \begin{tabular}{cc cc cc cc cc}
        \toprule
        \textbf{Acquisition} & \textbf{Train} & \multicolumn{2}{c}{\textbf{CIFAR-100 LB}} & \multicolumn{2}{c}{\textbf{CIFAR-100 HB}}
          & \multicolumn{2}{c}{\textbf{Imagewoof LB}}  & \multicolumn{2}{c}{\textbf{Imagewoof HB}} \\
        \cmidrule(lr){3-4} \cmidrule(lr){5-6} \cmidrule(lr){7-8} \cmidrule(lr){9-10}
        \textbf{ Func. $\mathcal{S}$} & \textbf{Method}
            & $A_{9}$ & $p$-value & $A_{9}$ & $p$-value & $A_{9}$ & $p$-value & $A_{9}$ & $p$-value \\
        \midrule
        \multirow{4}{*}{\shortstack{\textbf{Margin}\\SIGIR 1995}} & Dense          & $\mathbf{17.45 {\scriptstyle \pm 2.07}}$ & -- & $\mathbf{52.48 {\scriptstyle \pm 1.33}}$ & -- & $\mathbf{54.74 {\scriptstyle \pm 3.61}}$ & -- & $\mathbf{80.95 {\scriptstyle \pm 0.92}}$ & -- \\
                                & Rand Sparse    & $12.97 {\scriptstyle \pm 1.03}$ & {\small \xmark\ ($0.999$)} & $41.76 {\scriptstyle \pm 1.32}$ & {\small \xmark\ ($1.000$)} & $47.25 {\scriptstyle \pm 1.06}$ & {\small \xmark\ ($1.000$)} & $76.64 {\scriptstyle \pm 0.94}$ & {\small \xmark\ ($1.000$)} \\
                                & Rand Prune     & $11.70 {\scriptstyle \pm 1.18}$ & {\small \xmark\ ($1.000$)} & $17.46 {\scriptstyle \pm 1.46}$ & {\small \xmark\ ($1.000$)} & $43.64 {\scriptstyle \pm 1.54}$ & {\small \xmark\ ($1.000$)} & $64.04 {\scriptstyle \pm 3.21}$ & {\small \xmark\ ($1.000$)} \\
                                & \method (ours) & $\mathbf{17.52 {\scriptstyle \pm 1.80}}$ & {\small \cmark\ ($0.032$)} & $\mathbf{52.29 {\scriptstyle \pm 1.30}}$ & {\small \cmark\ ($0.014$)} & $\mathbf{55.46 {\scriptstyle \pm 3.44}}$ & {\small \cmark\ ($0.001$)} & $\mathbf{81.14 {\scriptstyle \pm 0.80}}$ & {\small \cmark\ ($0.001$)} \\
        \midrule
        \multirow{4}{*}{\shortstack{\textbf{CoreSet}\\ICLR 2018}} & Dense          & $\mathbf{17.73 {\scriptstyle \pm 2.09}}$ & -- & $\mathbf{51.37 {\scriptstyle \pm 1.23}}$ & -- & $\mathbf{55.15 {\scriptstyle \pm 4.29}}$ & -- & $\mathbf{79.82 {\scriptstyle \pm 0.82}}$ & -- \\
                                                         & Rand Sparse    & $13.78 {\scriptstyle \pm 0.71}$ & {\small \xmark\ ($0.997$)} & $41.91 {\scriptstyle \pm 1.13}$ & {\small \xmark\ ($1.000$)} & $46.49 {\scriptstyle \pm 1.83}$ & {\small \xmark\ ($1.000$)} & $75.34 {\scriptstyle \pm 1.00}$ & {\small \xmark\ ($1.000$)} \\
                                                         & Rand Prune     & $12.20 {\scriptstyle \pm 1.46}$ & {\small \xmark\ ($1.000$)} & $18.50 {\scriptstyle \pm 1.39}$ & {\small \xmark\ ($1.000$)} & $45.06 {\scriptstyle \pm 1.65}$ & {\small \xmark\ ($1.000$)} & $68.41 {\scriptstyle \pm 1.92}$ & {\small \xmark\ ($1.000$)} \\
                                                         & \method (ours) & $\mathbf{18.44 {\scriptstyle \pm 2.16}}$ & {\small \cmark\ ($0.001$)} & $\mathbf{50.52 {\scriptstyle \pm 1.03}}$ & {\small \xmark\ ($0.080$)} & $\mathbf{54.73 {\scriptstyle \pm 4.83}}$ & {\small \cmark\ ($0.001$)} & $\mathbf{79.07 {\scriptstyle \pm 1.14}}$ & {\small \cmark\ ($0.001$)} \\
        \midrule
        \multirow{4}{*}{\shortstack{\textbf{BADGE}\\ICLR 2020}} & Dense          & $\mathbf{18.01 {\scriptstyle \pm 2.44}}$ & -- & $\mathbf{50.89 {\scriptstyle \pm 1.35}}$ & -- & $\mathbf{54.79 {\scriptstyle \pm 3.63}}$ & -- & $\mathbf{78.68 {\scriptstyle \pm 1.07}}$ & -- \\
                                                       & Rand Sparse    & $13.83 {\scriptstyle \pm 0.81}$ & {\small \xmark\ ($0.995$)} & $41.63 {\scriptstyle \pm 1.12}$ & {\small \xmark\ ($1.000$)} & $47.80 {\scriptstyle \pm 1.41}$ & {\small \xmark\ ($1.000$)} & $75.43 {\scriptstyle \pm 0.63}$ & {\small \xmark\ ($1.000$)} \\
                                                       & Rand Prune     & $12.40 {\scriptstyle \pm 1.76}$ & {\small \xmark\ ($1.000$)} & $18.93 {\scriptstyle \pm 1.16}$ & {\small \xmark\ ($1.000$)} & $44.49 {\scriptstyle \pm 1.34}$ & {\small \xmark\ ($1.000$)} & $68.09 {\scriptstyle \pm 1.50}$ & {\small \xmark\ ($1.000$)} \\
                                                       & \method (ours) & $\mathbf{18.48 {\scriptstyle \pm 1.86}}$ & {\small \cmark\ ($0.001$)} & $\mathbf{49.94 {\scriptstyle \pm 1.34}}$ & {\small \xmark\ ($0.903$)} & $\mathbf{54.43 {\scriptstyle \pm 3.92}}$ & {\small \cmark\ ($0.002$)} & $\mathbf{78.45 {\scriptstyle \pm 1.24}}$ & {\small \cmark\ ($0.003$)} \\
        \midrule
        \multirow{4}{*}{\shortstack{\textbf{SAAL}\\PMLR 2023}} & Dense          & $\mathbf{17.76 {\scriptstyle \pm 1.91}}$ & -- & $\mathbf{51.74 {\scriptstyle \pm 1.17}}$ & -- & $\mathbf{54.83 {\scriptstyle \pm 3.19}}$ & -- & $\mathbf{80.24 {\scriptstyle \pm 1.08}}$ & -- \\
                                                      & Rand Sparse    & $13.10 {\scriptstyle \pm 0.95}$ & {\small \xmark\ ($0.997$)} & $41.76 {\scriptstyle \pm 1.34}$ & {\small \xmark\ ($1.000$)} & $47.12 {\scriptstyle \pm 1.30}$ & {\small \xmark\ ($1.000$)} & $76.45 {\scriptstyle \pm 0.81}$ & {\small \xmark\ ($1.000$)} \\
                                                      & Rand Prune     & $12.29 {\scriptstyle \pm 1.85}$ & {\small \xmark\ ($1.000$)} & $18.50 {\scriptstyle \pm 1.27}$ & {\small \xmark\ ($1.000$)} & $42.72 {\scriptstyle \pm 2.33}$ & {\small \xmark\ ($1.000$)} & $66.16 {\scriptstyle \pm 1.63}$ & {\small \xmark\ ($1.000$)} \\
                                                      & \method (ours) & $\mathbf{17.88 {\scriptstyle \pm 1.97}}$ & {\small \cmark\ ($0.005$)} & $\mathbf{51.36 {\scriptstyle \pm 1.32}}$ & {\small \xmark\ ($0.313$)} & $\mathbf{55.62 {\scriptstyle \pm 3.84}}$ & {\small \cmark\ ($0.001$)} & $\mathbf{81.01 {\scriptstyle \pm 1.27}}$ & {\small \cmark\ ($0.001$)} \\
        \midrule
        \multirow{4}{*}{\shortstack{\textbf{MaxHerding}\\ECCV 2024}} & Dense          & $\mathbf{18.72 {\scriptstyle \pm 2.01}}$ & -- & $\mathbf{51.89 {\scriptstyle \pm 0.93}}$ & -- & $\mathbf{55.74 {\scriptstyle \pm 3.91}}$ & -- & $\mathbf{80.33 {\scriptstyle \pm 1.07}}$ & -- \\
                                                            & Rand Sparse    & $13.92 {\scriptstyle \pm 1.00}$ & {\small \xmark\ ($0.999$)} & $41.88 {\scriptstyle \pm 1.37}$ & {\small \xmark\ ($1.000$)} & $47.76 {\scriptstyle \pm 1.64}$ & {\small \xmark\ ($1.000$)} & $76.41 {\scriptstyle \pm 1.02}$ & {\small \xmark\ ($1.000$)} \\
                                                            & Rand Prune     & $12.41 {\scriptstyle \pm 1.73}$ & {\small \xmark\ ($1.000$)} & $19.03 {\scriptstyle \pm 1.36}$ & {\small \xmark\ ($1.000$)} & $42.88 {\scriptstyle \pm 1.76}$ & {\small \xmark\ ($1.000$)} & $65.38 {\scriptstyle \pm 2.60}$ & {\small \xmark\ ($1.000$)} \\
                                                            & \method (ours) & $\mathbf{18.89 {\scriptstyle \pm 1.88}}$ & {\small \cmark\ ($0.005$)} & $\mathbf{51.39 {\scriptstyle \pm 1.16}}$ & {\small \xmark\ ($0.116$)} & $\mathbf{55.16 {\scriptstyle \pm 4.13}}$ & {\small \cmark\ ($0.003$)} & $\mathbf{80.55 {\scriptstyle \pm 1.27}}$ & {\small \cmark\ ($0.001$)} \\
        \midrule
        \multirow{4}{*}{\shortstack{\textbf{UHerding}\\ICLR 2025}} & Dense          & $\mathbf{17.70 {\scriptstyle \pm 2.20}}$ & -- & $\mathbf{52.28 {\scriptstyle \pm 1.38}}$ & -- & $\mathbf{54.70 {\scriptstyle \pm 2.75}}$ & -- & $\mathbf{80.46 {\scriptstyle \pm 1.14}}$ & -- \\
                                                          & Rand Sparse    & $13.87 {\scriptstyle \pm 0.81}$ & {\small \xmark\ ($0.993$)} & $41.81 {\scriptstyle \pm 1.16}$ & {\small \xmark\ ($1.000$)} & $47.45 {\scriptstyle \pm 1.71}$ & {\small \xmark\ ($1.000$)} & $77.06 {\scriptstyle \pm 1.17}$ & {\small \xmark\ ($1.000$)} \\
                                                          & Rand Prune     & $12.27 {\scriptstyle \pm 1.18}$ & {\small \xmark\ ($1.000$)} & $17.68 {\scriptstyle \pm 1.14}$ & {\small \xmark\ ($1.000$)} & $42.40 {\scriptstyle \pm 2.05}$ & {\small \xmark\ ($1.000$)} & $63.45 {\scriptstyle \pm 2.28}$ & {\small \xmark\ ($1.000$)} \\
                                                          & \method (ours) & $\mathbf{18.19 {\scriptstyle \pm 1.86}}$ & {\small \cmark\ ($0.001$)} & $\mathbf{51.86 {\scriptstyle \pm 1.39}}$ & {\small \cmark\ ($0.014$)} & $\mathbf{55.52 {\scriptstyle \pm 3.34}}$ & {\small \cmark\ ($0.001$)} & $\mathbf{81.07 {\scriptstyle \pm 1.11}}$ & {\small \cmark\ ($0.001$)} \\
        \bottomrule
    \end{tabular}}%
\end{table}

\section{Experiments}\label{sec:eval}
We evaluate \method on four image classification datasets: CIFAR-100~\citep{cifar}, Imagewoof~\citep{Howard_Imagewoof_2019}, Tiny-ImageNet~\citep{tiny_imagenet}, and Places365~\citep{zhou2017places}. These datasets cover training from scratch at multiple input resolutions as well as active finetuning from ImageNet-pretrained checkpoints. For CIFAR-100, Imagewoof, and Tiny-ImageNet a ResNet-18~\citep{resnet}, we pretrain a self-supervised backbone with SimCLR~\citep{chen2020simple} on the whole unlabeled pool \emph{before} any annotation is drawn, reflecting the now-standard practice of initializing DAL pipelines from a representation learned on the unlabeled data rather than from random weights~\citep{gashi2025deep}, a convention adopted for robust active learning evaluation~\citep{lueth2023navigating} rather than a requirement of \method itself. We also adapt the stem of the model per input resolution to avoid excessive early down-sampling (Appendix~\ref{app:resnet_adaptations}). All models are trained with AdamW~\citep{loshchilov2018decoupled} and a cosine learning-rate schedule with linear warmup. Convergence is considered on the first epoch that reaches a fixed high training-accuracy threshold on the current labeled set, following~\citet{ash2019deep} early stopping criterion. At each DAL iteration, we re-initialize the model to the pretrained checkpoint to prevent warm-start interference~\citep{ash2020warm}. Normalization layers and the classification head are excluded from pruning. Full per-dataset hyperparameters, including optimizer settings, schedule, budgets and pruning configuration are reported in Table~\ref{tab:hyperparams}. 

\newpage
\begin{table}
    \centering
    \caption{\emph{Longer DAL cycle}. Classification accuracy [\%] (mean $\pm$ std over \num{15} seeds) on Tiny-ImageNet, starting from \num{1000} stratified samples with a per-iteration budget of \num{200} over \num{20} iterations. $p$-value of one-sided Wilcoxon test for non-inferiority against Dense on AULC; \cmark\ if $p<0.05$ (margin $\epsilon=0.005$), \xmark\ otherwise.
    }
    \label{tab:longer_al_cycle}
    
    \resizebox{.95\linewidth}{!}{%
    \begin{tabular}{ccccccccc}
        \toprule
        \textbf{Acquisition} & \textbf{Train} & \multicolumn{7}{c}{\textbf{Tiny-ImageNet}} \\ \cmidrule(lr){3-9}
        \textbf{Func. $\mathcal{S}$} & \textbf{Method}
            & $A_0$ & $A_4$ & $A_8$ & $A_{12}$ & $A_{16}$ & $A_{20}$ & {$p$-value} \\
        \midrule
        \multirow{4}{*}{\shortstack{\textbf{Margin}\\SIGIR 1995}}
            & Dense                    & $\mathbf{10.56 {\scriptstyle \pm 0.49}}$ & $\mathbf{14.07 {\scriptstyle \pm 0.65}}$ & $\mathbf{16.70 {\scriptstyle \pm 0.62}}$ & $\mathbf{19.28 {\scriptstyle \pm 0.56}}$ & $\mathbf{21.15 {\scriptstyle \pm 0.70}}$ & $\mathbf{22.77 {\scriptstyle \pm 0.72}}$ & -- \\
            & Rand Sparse              & $8.27 {\scriptstyle \pm 0.43}$  & $6.86 {\scriptstyle \pm 0.42}$  & $8.59 {\scriptstyle \pm 0.48}$  & $10.28 {\scriptstyle \pm 0.56}$ & $11.77 {\scriptstyle \pm 0.37}$ & $13.32 {\scriptstyle \pm 0.60}$ & {\small \xmark\ ($1.00$)} \\
            & Rand Prune               & $9.42 {\scriptstyle \pm 0.52}$  & $7.62 {\scriptstyle \pm 0.48}$  & $9.09 {\scriptstyle \pm 0.53}$  & $9.67 {\scriptstyle \pm 0.77}$  & $11.02 {\scriptstyle \pm 0.67}$ & $10.64 {\scriptstyle \pm 0.77}$ & {\small \xmark\ ($1.00$)} \\
            & \method (ours)              & $\mathbf{11.27 {\scriptstyle \pm 0.54}}$ & $\mathbf{14.10 {\scriptstyle \pm 0.65}}$ & $\mathbf{16.87 {\scriptstyle \pm 0.64}}$ & $\mathbf{18.83 {\scriptstyle \pm 0.64}}$ & $\mathbf{20.71 {\scriptstyle \pm 0.63}}$ & $\mathbf{22.22 {\scriptstyle \pm 0.54}}$ & {\small \cmark\ ($3e^{-5}$) } \\
        \midrule
        \multirow{4}{*}{\shortstack{\textbf{CoreSet}\\ICLR 2018}}
            & Dense                    & $\mathbf{10.56 {\scriptstyle \pm 0.49}}$ & $\mathbf{13.97 {\scriptstyle \pm 0.51}}$ & $\mathbf{16.77 {\scriptstyle \pm 0.79}}$ & $\mathbf{19.32 {\scriptstyle \pm 0.61}}$ & $\mathbf{20.84 {\scriptstyle \pm 0.79}}$ & $\mathbf{22.69 {\scriptstyle \pm 0.68}}$ & -- \\
            & Rand Sparse              & $8.27 {\scriptstyle \pm 0.43}$  & $6.65 {\scriptstyle \pm 0.34}$  & $8.26 {\scriptstyle \pm 0.45}$  & $9.87 {\scriptstyle \pm 0.41}$  & $11.12 {\scriptstyle \pm 0.37}$ & $12.32 {\scriptstyle \pm 0.44}$ & {\small \xmark\ ($1.00$)} \\
            & Rand Prune               & $9.42 {\scriptstyle \pm 0.52}$  & $7.87 {\scriptstyle \pm 0.35}$  & $9.24 {\scriptstyle \pm 0.37}$  & $10.29 {\scriptstyle \pm 0.57}$ & $11.57 {\scriptstyle \pm 0.66}$ & $11.71 {\scriptstyle \pm 0.81}$ & {\small \xmark\ ($1.00$)} \\
            & \method (ours)              & $\mathbf{11.27 {\scriptstyle \pm 0.54}}$ & $\mathbf{14.32 {\scriptstyle \pm 0.65}}$ & $\mathbf{17.20 {\scriptstyle \pm 0.64}}$ & $\mathbf{18.93 {\scriptstyle \pm 0.59}}$ & $\mathbf{20.70 {\scriptstyle \pm 0.49}}$ & $\mathbf{22.20 {\scriptstyle \pm 0.44}}$ & {\small \cmark\ ($3e^{-5}$) } \\
        \midrule
        \multirow{4}{*}{\shortstack{\textbf{BADGE}\\ICLR 2020}}
            & Dense                    & $\mathbf{10.56 {\scriptstyle \pm 0.49}}$ & $\mathbf{14.42 {\scriptstyle \pm 0.40}}$ & $\mathbf{17.43 {\scriptstyle \pm 0.63}}$ & $\mathbf{19.59 {\scriptstyle \pm 0.53}}$ & $\mathbf{21.60 {\scriptstyle \pm 0.70}}$ & $\mathbf{23.50 {\scriptstyle \pm 0.72}}$ & -- \\
            & Rand Sparse              & $8.27 {\scriptstyle \pm 0.43}$  & $7.10 {\scriptstyle \pm 0.36}$  & $8.84 {\scriptstyle \pm 0.42}$  & $10.85 {\scriptstyle \pm 0.30}$ & $12.36 {\scriptstyle \pm 0.30}$ & $13.94 {\scriptstyle \pm 0.52}$ & {\small \xmark\ ($1.00$)} \\
            & Rand Prune               & $9.42 {\scriptstyle \pm 0.52}$  & $8.00 {\scriptstyle \pm 0.42}$  & $9.76 {\scriptstyle \pm 0.31}$  & $10.42 {\scriptstyle \pm 0.56}$ & $12.15 {\scriptstyle \pm 0.62}$ & $11.83 {\scriptstyle \pm 0.76}$ & {\small \xmark\ ($1.00$)} \\
            & \method (ours)              & $\mathbf{11.27 {\scriptstyle \pm 0.54}}$ & $\mathbf{14.44 {\scriptstyle \pm 0.46}}$ & $\mathbf{17.42 {\scriptstyle \pm 0.64}}$ & $\mathbf{19.16 {\scriptstyle \pm 0.55}}$ & $\mathbf{21.47 {\scriptstyle \pm 0.62}}$ & $\mathbf{22.81 {\scriptstyle \pm 0.50}}$ & {\small \cmark\ ($3e^{-5}$) } \\
        \bottomrule
    \end{tabular}}%
\end{table}

We compare \method against three training baselines (further pruning baselines are investigated in Appendix~\ref{app:pai_rigl}):

\begin{itemize}
    \item \textbf{Dense.} The standard DAL pipeline without any pruning steps (finetuning phase still applies). The model trains at full capacity in every iteration and serves as the target upper-bound reference.
    \item \textbf{Rand Sparse.} The initial model is pruned at the same sparsity $s$ as \method. Training proceeds in a single stage without the expand-and-contract cycle or the second-stage finetuning. This baseline evaluates whether merely operating at reduced capacity, without structured mask discovery, is sufficient.
    \item \textbf{Rand Prune.} The same pipeline as \method is retained, including reactivation $\mathcal{R}$ and the finetuning phase, but magnitude-based pruning is replaced by random pruning.
\end{itemize}

\subsection{Main experiments}

\begin{wrapfigure}{r}{0.5\linewidth}
    \vspace{-1.25em}
    \centering
    \includegraphics[width=\linewidth]{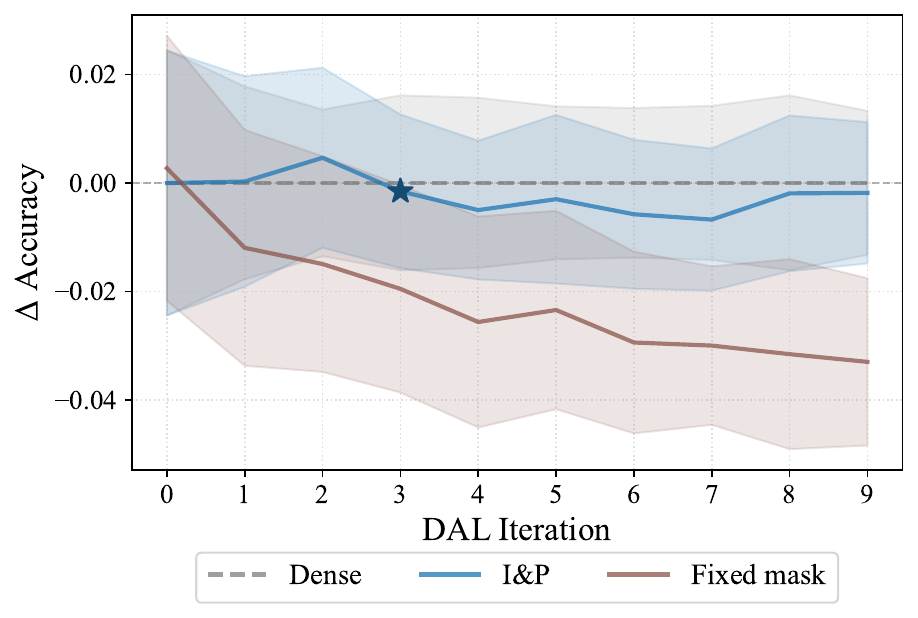}
    \caption{Fixed transfer mask vs.\ \method on \mbox{CIFAR-100~HB} (Margin). Relative accuracy is reported w.r.t.\ Dense over 5 seeds. The $\bigstar$ marks the iteration where \method reaches $\approx\!95\%$ sparsity, which corresponds to the sparsity of \emph{fixed mask} at all iterations.
    }
    \label{fig:transfer_mask}
    \vspace{-1em}
\end{wrapfigure}

The choice of acquisition function affects the distribution coverage of the model through the choice of labeled batches. Therefore, we aim to elucidate how this choice interacts with pruning. We report final test accuracy $A_{9}$ on \mbox{CIFAR-100} and Imagewoof across six acquisition functions within a low (LB) and high (HB) annotation budget regime. Specifically, LB sets both the initial pool $|D_L^{(0)}|$ and the per-round budget $B$ between 1-5 samples per class, while HB scales them an order of magnitude higher to 10-50 samples per class. We additionally perform a one-sided Wilcoxon signed-rank (non-inferiority) test on the AULC~\citep{werner2024a}, with hypothesis \mbox{$H_1:\mathrm{AULC}_{A} - \mathrm{AULC}_{B} < \epsilon $}. If rejected, method $B$ is unlikely to fall below $\epsilon=0.005$ AULC compared to method $A$ (see Appendix~\ref{app:wilcoxon} for details). In our experiments, Dense is the reference method $A$.

Results are presented in \Cref{tab:final_acc_results}, which show how \method consistently matches Dense across all settings, confirming that the DAL retraining loop produces winning tickets without a dedicated, extra ticket-search cost. Furthermore, the training method is agnostic to the choice of acquisition function: no function exhibits a systematic accuracy loss under pruning. In Appendix~\ref{app:mask_trajectory}, we explore whether structural similarities emerge from different acquisition functions within the CIFAR HB setting. Appendix~\ref{app:learning_curves} contains detailed visualizations over the DAL cycle. Noticeably, both random baselines fall considerably below Dense, emphasizing that no trivial sparse solution is available.

The preceding experiments use a moderate number ($A\!=\!9$) of DAL iterations, which might hide longer cycle performance trends. Therefore, we also check a longer iteration cycle ($A\!=\!20$) on Tiny-ImageNet. The results are summarized in \Cref{tab:longer_al_cycle}. The longer cycle exposes no cumulative pruning penalty, indicating that \method remains viable as the number of DAL rounds grows, which is validated by the statistical test.

\minisection{Cross-iteration mask transferability.}
\citet{zhang2021efficient} have shown that masks can transfer across related settings, which questions the need for per-iteration re-discovery. To address this, we apply the standard (non-DAL) IMP on the initial labeled pool \mbox{$D^{(0)}_{L}$} until only $(1 - s)$ active parameters remain. This \emph{fixed mask} is applied throughout the remaining DAL iterations. \Cref{fig:transfer_mask} compares this strategy against \method on \mbox{CIFAR-100~HB} (Margin), and shows how both achieve similar initial accuracy. However, the \emph{fixed mask} degrades steadily as more data is acquired.
In contrast, \method refines the mask at every DAL iteration maintaining near-Dense accuracy for the whole cycle.

\subsection{Computational costs analysis}
\begin{wrapfigure}{R}{0.5\textwidth}
    \centering
    \captionof{table}{Training (T) and finetuning (FT) epochs per DAL iteration on Tiny-ImageNet, averaged across acquisition functions, DAL rounds, and seeds. \emph{Conv.}\ is the fraction of trainings that re-reach the train-acc threshold before patience budget; values below $100\%$ indicate runs terminated at the patience ceiling. 
    }
    \label{tab:stage2_cost}
    \begin{tabular}{lccc}
        \toprule
        \textbf{Train} & \textbf{T epochs} & \textbf{FT epochs} & \textbf{Conv. (\%)} \\
        \midrule
        Dense       & $63.1 {\scriptstyle \pm 16.6}$ & $3.3 {\scriptstyle \pm 4.3}$  & $100.0$  \\
        \method (ours) & $56.5 {\scriptstyle \pm 19.6}$ & $3.8 {\scriptstyle \pm 2.3}$  & $100.0$ \\
        Rand Prune  & $80.7 {\scriptstyle \pm 19.6}$ & $29.8 {\scriptstyle \pm 4.0}$ & $9.5$   \\
        \bottomrule
    \end{tabular}
    
    \vspace{2.18em}

    \includegraphics[width=\linewidth]{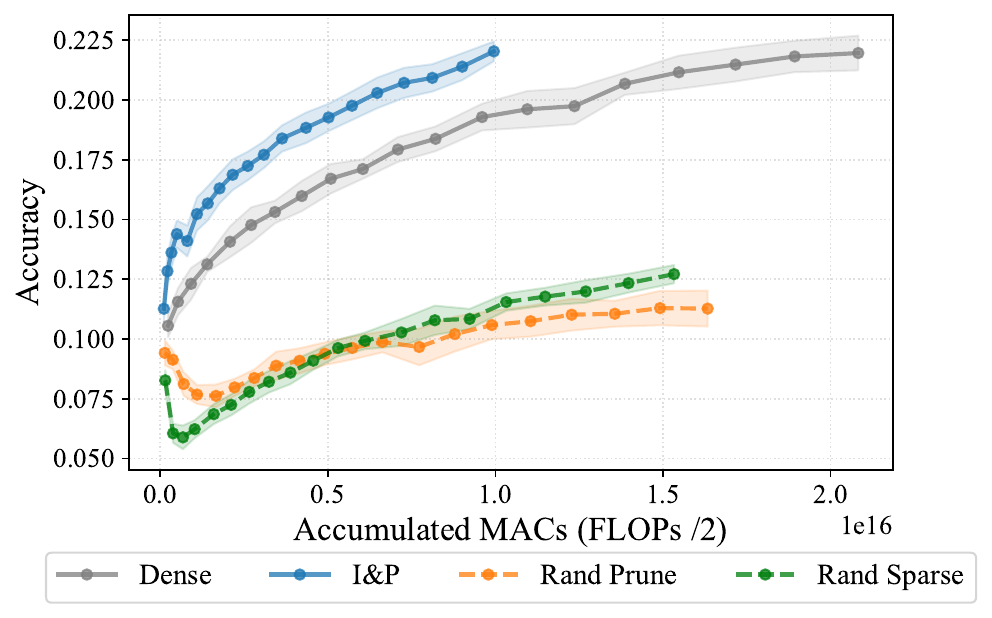}
    \captionof{figure}{\emph{Mean Accuracy with one std as a function of cumulative MACs} calculated over \num{15} seeds with $\mathcal{S}$ = Margin  on Tiny-ImageNet. Each marker corresponds to one DAL iteration. Curves that stop at lower cumulative MAC values indicate a more efficient pipeline at comparable accuracy.}
    \label{fig:accuracy_vs_flops}
    \vspace{-4em}
\end{wrapfigure}

\minisection{Finetuning cost and mask quality.}
The purpose of the finetuning stage is to compensate for any catastrophic degradation in generalization induced by pruning, so that the acquisition function can exploit the pruned model in the subsequent selection step. For this to be viable, recovery must be achievable at negligible compute relative to main training; otherwise the computational benefits of sparse models are offset by a prohibitive retraining cost. We analyze finetuning costs as a direct probe of mask quality: a valid pruning signal should permit rapid re-convergence, whereas an uninformative mask should not. \Cref{tab:stage2_cost} reports finetuning epochs and convergence rate on Tiny-ImageNet. \method re-converges within a few epochs on every run, indistinguishable from Dense finetuning. In contrast, Rand Prune fails to recover on more than $90\%$ of runs and is terminated at the patience ceiling, confirming that random masks at the same sparsity do not preserve a trainable subnetwork. The same pattern holds for CIFAR-100, Imagewoof, and Places365. The consequences of omitting the finetune phase have been explored in Appendix~\ref{app:stage2_cost}.

\minisection{FLOPs Approximation.}
All computational savings reported in this work are \emph{theoretical}; we do not measure wall-clock time on specific hardware (see \Cref{sec:limitations} for discussion). We estimate multiply-accumulate operations (MACs) in a \emph{mask-aware, per-layer} fashion rather than applying a single global scaling factor~\citep{hoefler2021sparsity}, because the contribution of each pruned parameter depends on the layer type and the spatial dimensions of the intermediate feature maps at that stage. For MACs calculation details and a general scaling of acquisition function computation see Appendix~\ref{app:flops_details} and \ref{app:flops_scaling}, respectively.

\Cref{fig:accuracy_vs_flops} visualizes the cumulative computational cost paired with achieved test accuracy  on Tiny-ImageNet. \method reaches the same test accuracy as the dense baseline while spending roughly half the computations across the full $20$-iteration DAL cycle. Both random baselines land in between due to more training epochs from the exhausted patience $\varphi$ budget (see~\Cref{tab:stage2_cost}), inflating their training cost to roughly $50\%$ above that of \method, while converging to substantially lower accuracy. The same accuracy vs MACs trend holds for CIFAR-100, Imagewoof and Places365 and across acquisition functions (see Appendix~\ref{app:accuracy_vs_flops}).

\begin{table}
    \centering
    \caption{\emph{Pretrained features survive pruning under active finetuning}. Classification accuracy [\%] (mean $\pm$ std over \num{3} seeds) on Places365 with ImageNet pretrained backbone and Margin acquisition. Initial label pool of \num{5475} class-balanced samples with an annotation budget of \num{3650} over \num{9} iterations.
    }
    \label{tab:arch_viability}
    \resizebox{\linewidth}{!}{%
    \begin{tabular}{cccccccc}
        \toprule
        \textbf{Architecture} & \textbf{Train}
            & $A_0$ & $A_1$ & $A_3$ & $A_5$ & $A_7$ & $A_9$ \\
        \midrule
        \multirow{3}{*}{\shortstack{ResNet-50 \\ (sup.)}}
            & Dense       & $\mathbf{21.49 {\scriptstyle\pm 0.28}}$ & $\mathbf{24.96 {\scriptstyle\pm 0.35}}$ & $\mathbf{28.70 {\scriptstyle\pm 0.34}}$ & $\mathbf{31.23 {\scriptstyle\pm 0.30}}$ & $\mathbf{32.78 {\scriptstyle\pm 0.17}}$ & $\mathbf{34.20 {\scriptstyle\pm 0.09}}$ \\
            & Rand Sparse & $19.94 {\scriptstyle\pm 0.36}$ & $22.72 {\scriptstyle\pm 0.29}$ & $25.40 {\scriptstyle\pm 0.17}$ & $26.24 {\scriptstyle\pm 0.20}$ & $24.43 {\scriptstyle\pm 0.08}$ & $26.28 {\scriptstyle\pm 0.63}$ \\
            & \method (ours) & $\mathbf{21.89 {\scriptstyle\pm 0.07}}$ & $\mathbf{25.48 {\scriptstyle\pm 0.33}}$ & $\mathbf{28.70 {\scriptstyle\pm 0.31}}$ & $30.36 {\scriptstyle\pm 0.12}$ & $32.04 {\scriptstyle\pm 0.13}$ & $33.16 {\scriptstyle\pm 0.15}$ \\
        \midrule
        \multirow{3}{*}{\shortstack{ConvNext v2 \\ Tiny (ssl.)}}
            & Dense       & $\mathbf{18.96 {\scriptstyle\pm 0.30}}$ & $\mathbf{22.92 {\scriptstyle\pm 0.19}}$ & $\mathbf{26.21 {\scriptstyle\pm 0.77}}$ & $\mathbf{27.61 {\scriptstyle\pm 1.21}}$ & $\mathbf{27.16 {\scriptstyle\pm 4.03}}$ & $\mathbf{30.44 {\scriptstyle\pm 0.52}}$ \\
            & Rand Sparse & $8.81 {\scriptstyle\pm 0.70}$  & $10.69 {\scriptstyle\pm 0.65}$ & $14.93 {\scriptstyle\pm 0.78}$ & $17.64 {\scriptstyle\pm 0.82}$ & $19.00 {\scriptstyle\pm 0.11}$ & $20.93 {\scriptstyle\pm 0.40}$ \\
            & \method (ours) & $18.38 {\scriptstyle\pm 0.20}$ & $\mathbf{22.89 {\scriptstyle\pm 0.10}}$ & $\mathbf{26.31 {\scriptstyle\pm 0.24}}$ & $\mathbf{27.82 {\scriptstyle\pm 0.10}}$ & $\mathbf{28.63 {\scriptstyle\pm 0.27}}$ & $\mathbf{30.00 {\scriptstyle\pm 0.23}}$ \\
        \midrule
        \multirow{3}{*}{\shortstack{DeiT-Small \\ (sup.)}}
            & Dense       & $\mathbf{31.99 {\scriptstyle\pm 0.48}}$ & $\mathbf{33.99 {\scriptstyle\pm 0.26}}$ & $\mathbf{35.56 {\scriptstyle\pm 0.08}}$ & $\mathbf{34.85 {\scriptstyle\pm 0.47}}$ & $\mathbf{34.82 {\scriptstyle\pm 0.26}}$ & $\mathbf{35.61 {\scriptstyle\pm 0.30}}$ \\
            & Rand Sparse & $27.32 {\scriptstyle\pm 0.31}$ & $26.92 {\scriptstyle\pm 0.44}$ & $27.27 {\scriptstyle\pm 0.30}$ & $27.29 {\scriptstyle\pm 0.51}$ & $27.68 {\scriptstyle\pm 0.31}$ & $28.62 {\scriptstyle\pm 0.09}$ \\
            & \method (ours) & $\mathbf{31.92 {\scriptstyle\pm 0.21}}$ & $33.55 {\scriptstyle\pm 0.24}$ & $33.44 {\scriptstyle\pm 0.28}$ & $33.84 {\scriptstyle\pm 0.09}$ & $34.30 {\scriptstyle\pm 0.07}$ & $34.94 {\scriptstyle\pm 0.04}$ \\
        \midrule
        \multirow{3}{*}{\shortstack{DINO ViT-S \\ (ssl.)}}
            & Dense       & $\mathbf{17.72 {\scriptstyle\pm 0.43}}$ & $\mathbf{21.53 {\scriptstyle\pm 0.34}}$ & $\mathbf{25.21 {\scriptstyle\pm 0.30}}$ & $\mathbf{27.41 {\scriptstyle\pm 0.17}}$ & $\mathbf{27.20 {\scriptstyle\pm 1.24}}$ & $\mathbf{29.38 {\scriptstyle\pm 1.08}}$ \\
            & Rand Sparse & $12.80 {\scriptstyle\pm 0.20}$ & $15.34 {\scriptstyle\pm 0.26}$ & $17.94 {\scriptstyle\pm 0.08}$ & $18.76 {\scriptstyle\pm 0.78}$ & $19.65 {\scriptstyle\pm 0.26}$ & $21.63 {\scriptstyle\pm 0.33}$ \\
            & \method (ours) & $\mathbf{17.85 {\scriptstyle\pm 0.27}}$ & $\mathbf{21.52 {\scriptstyle\pm 0.29}}$ & $24.29 {\scriptstyle\pm 0.29}$ & $26.32 {\scriptstyle\pm 0.29}$ & $\mathbf{27.64 {\scriptstyle\pm 0.03}}$ & $27.46 {\scriptstyle\pm 0.83}$ \\
        \bottomrule
    \end{tabular}}%
\end{table}

\subsection{Active Finetuning}
To test whether \method is suitable for transfer learning, we evaluate an active finetuning scenario~\citep{xie2023active} on Places365. Model checkpoints pretrained on ImageNet are adapted to the target domain. This setting is particularly demanding for pruning: the pretrained features posses strong generalization potential, and aggressive weight removal risk catastrophic forgetting of the learned representations. We evaluate on four architectures and two training paradigms: ResNet-50~\citep{resnet} and DeiT-Small~\citep{deit} with supervised ImageNet pretraining, and ConvNext~v2 Tiny~\citep{convnextv2} and DINO ViT-S~\citep{caron2021emerging} with self-supervised pretraining. LTH discovery with magnitude pruning is significantly more difficult for ViTs than CNNs~\citep{liu2024survey}. To stabilize the discovery we keep the first 3 ViT blocks dense to stabilize features as they are especially sensitive for modification~\citep{deit}. Other specialized LTH pruning signals for ViTs exist~\citep{ijcai2023p153} and could be adopted seemingly.

On Places365 active finetuning (\Cref{tab:arch_viability}), \method stays within an absolute margin of $\epsilon \approx 1\%$ of Dense across all four backbones, although the difference is larger than in the other settings. The residual gap is most visible on DeiT-Small and DINO ViT-S, yet \method remains decisively above Rand~Sparse on every architecture, confirming that viability stems from an appropriate choice of $\mathcal{C}$ and $p_{\text{c}}$ rather than reduced capacity $s$ alone. Seed variance is elevated throughout—most notably for Dense ConvNext~v2, where per-seed deviations reach several percentage points—reflecting the limited budget of \num{3} seeds and cautioning against over-interpreting individual endpoint differences. Supervised ImageNet pretraining yields a more favourable starting point for the setting than the self-supervision.

\section{Discussion and Limitations}\label{sec:limitations}

\minisection{Practical realizability of computational savings.}
All FLOP reductions reported in this work are theoretical estimates derived from the sparsity of the pruning mask. On current mainstream GPU accelerators, unstructured sparsity does not translate directly into proportional wall-clock speedups, because dense matrix-multiplication kernels cannot exploit arbitrary zero patterns efficiently~\citep{hoefler2021sparsity}. Recent hardware generations have begun to close this gap: NVIDIA's Ampere and Hopper architectures provide native support for 2:4 semi-structured sparsity, achieving near $2{\times}$ throughput for conforming patterns~\citep{mishra2021accelerating}. While the masks produced by \method are fully unstructured, post-hoc conversion to 2:4 patterns or block-sparse layouts is an active area of research and could be applied as a deployment step. On CPU and edge devices, where inference relies on sparse linear-algebra libraries, the theoretical savings are directly realizable. We therefore view the FLOPs analysis as a proxy for the efficiency frontier as sparse hardware and software support continues to mature.

\minisection{Empirical scope and domain limitations.}
Our evaluation is limited to image classification tasks. Although the structural correspondence between the DAL iterative training loop and IMP is task-agnostic in principle, the lottery ticket hypothesis itself is an empirical observation: a winning ticket is not guaranteed to exist for every architecture, dataset, or sparsity level. Whether the favorable ticket discovery we observe generalizes to other domains (object detection, natural language processing, tabular data) remains an open question that warrants dedicated investigation. Similarly, our experiments focus on unstructured magnitude pruning; extending the framework to structured pruning criteria could yield more hardware-friendly sparse models but may require different reactivation strategies and sparsity schedules.

\minisection{Efficient selection of target sparsity.}
Most pruning methods require a target sparsity $s$ as a core hyperparameter~\citep{cheng2024survey}. IMP-style approaches additionally specify a per-round pruning rate $p_{\text{c}}$ that controls the granularity of iterative mask refinement. The finetuning phase of \method turns sparsity selection into a possible
post-hoc search: from the converged model at the end of main training, any candidate sparsity $s$ requires only applying a new mask and running the short finetuning phase with a reset learning rate
schedule~\citep{Renda2020Comparing}, while the expensive main training is shared across all candidates.

\minisection{Synergies between DAL and LTH.}
Beyond the structural correspondence exploited by our method, several independent best practices in the DAL and LTH communities converge in ways that may further facilitate ticket discovery within an active learning loop. Distributional acquisition functions~\citep{bae2024generalized}, which prioritize representative samples and have strong performance in initial DAL iterations~\citep{yehuda2022active}, produce training sets that resemble the easy-to-learn subsets found by~\citet{lth_data_diet} to improve lottery ticket discovery stability. Besides test accuracy, model quality has several facets that matter directly for acquisition. Winning tickets have been reported to exhibit improved uncertainty calibration~\citep{arora2023quantifying} and stronger out-of-distribution sensitivity than their dense counterparts~\citep{chen2022can}. This translates into cleaner signals for uncertainty-based criteria such as Margin or BADGE. The same works also point to more faithful learned representations, a property that should benefit coreset methods whose selections depend on feature-space geometry. Using the pruned model, rather than the dense checkpoint, to score the pool in the subsequent DAL round may thus improve the informativeness of acquired batches at no additional training cost.

\section{Conclusion}\label{sec:conclusion}
We presented \methodlong, a method that exploits the structural correspondence between the iterative retraining loop of pool-based active learning and iterative magnitude pruning. By appending a single pruning step to each DAL iteration, the method is able to uncover \emph{winning tickets} at nearly no additional training cost and without modification to the training procedure or acquisition function. Across four image classification datasets, six acquisition functions, and both CNN and Transformer architectures, sparse models produced by \method consistently match the accuracy of their dense counterparts while reducing FLOPs for both training and acquisition function sampling.
In particular, we observe that a fixed mask discovered through IMP early in the DAL cycle degrades as the labeled set grows, underscoring the need for expansion and contraction in each iteration of \method to keep the sparsity pattern aligned with the evolving data distribution. Together, these results provide the first evidence that winning tickets can be discovered even under a non-stationary data regime.

\section*{Acknowledgments}
This research was partially funded by the Austrian Science Fund (FWF) 10.55776/COE12. The authors sincerely thank their colleagues at Know-Center and the members of the CLAR group at the Institute of Visual Computing for the many insightful discussions, helpful feedback, and support that contributed to this work. We further acknowledge the Institute Human-Centered Computing for kindly providing essential compute resources and the Graz Center for Machine Learning (GraML) for providing an inspiring interdisciplinary research environment.

\bibliography{collas2026_conference}
\bibliographystyle{collas2026_conference}

\newpage
\appendix
\section{Appendix}

\subsection{Method Algorithm}\label{app:Algorithm}
\Cref{alg:method} gives the full \method procedure in pseudo-code, in line with the four stages of the pipeline overview in \Cref{fig:pipeline}. Each DAL iteration $(j)$ performs:
\begin{enumerate}
    \item \emph{Masked initialization} (lines~5-6): reactivate a fraction $p_{\text{react}}$ of previously pruned parameters via $\mathcal{R}$ and reload $\vtheta_{\text{init}}$ under the relaxed mask $M^{(j)}_{\text{init}}$.
    \item \emph{Model training} (line~7): an new model on the current labeled pool $D_L^{(j)}$ until the convergence criterion is met, yielding $\vtheta_K^{(j)}$.
    \item \emph{Prune and finetune} (lines~8--9): update the mask via $\mathcal{C}$ by zeroing the lowest-magnitude $p_{\text{c}}$ percentage of entries subject to the minimum active fraction $s$, and run a short finetuning phase of at most $\varphi$ epochs to compensate for the function drift induced by pruning.
    \item \emph{Acquisition function} (lines~15--18): score the unlabeled pool $D_U^{(j)}$ with the sparse finetuned model through $\mathcal{S}$, query the oracle, and update the labeled and unlabeled partitions accordingly.
\end{enumerate}
The optional rewind block (lines~11--13) allows $\vtheta_{\text{init}}$ to be replaced by an early training checkpoint $\vtheta_r^{(0)}$, recovering late rewinding proposed by~\citet{frankle2019stabilizing} as a special case.

\begin{algorithm}
  \caption{\methodlong}
  \label{alg:method}
  \begin{spacing}{1.35}
  \begin{algorithmic}[1]
    \STATE {\bfseries Input:}  Initial $D_L^{(0)}$ and $D_U^{(0)}$, initialization $\theta_{\text{init}}$, 
    DAL iterations $A$, target sparsity $s$, Reactivation function $\mathcal{R}$, Pruning function $\mathcal{C}$, Acquisition function $\mathcal{S}$, pruning rate $p_{\text{c}}$, reactivation rate $p_{\text{r}}$
    \STATE {\bfseries Optional:} rewind epoch $r = 0$; $M^{(0)} = \textbf{1}$
    \STATE
    \FOR{$j = 0$ to $A$}
    \STATE $M^{(j)} \leftarrow \mathcal{R}\left(M^{(j)} \, ; \, p_{\text{r}}\right)$
    \STATE $\vtheta^{(j)} \leftarrow\vtheta_{\text{init}} \odot M^{(j)}$    
    \STATE $\vtheta^{(j)}_K \xleftarrow{train} f\left(\vx; \vtheta^{(j)}\right) \ \text{on} \ D_L^{(j)}$
    \STATE $M^{(j+1)} \leftarrow \mathcal{C}\left(M^{(j)} \, ; \,  \vtheta_K^{(j)} \, ; \, p_{\text{c}} \, ; \,  s\right)$
    \STATE $\tilde{\vtheta}^{(j)}_{K+\varphi} \xleftarrow{finetune} f\left(\vx; \vtheta_K^{(j)} \odot M^{(j+1)}\right)$ \ \text{on} \ $D_L^{(j)}$

    \STATE
    \IF{$j=0$ \AND $r > 0$}
      \STATE $\vtheta_{\text{init}} \leftarrow \vtheta_r^{(0)}$
    \ENDIF
    \STATE
    \IF{$j \neq A$}
        \STATE $D_A^{(j)} \leftarrow \mathcal{S}\left(\tilde{\vtheta}_{K+\varphi}^{(j)} \, ;\,  D_U^{(j)} \right)$
        \STATE $D_L^{(j+1)} = D_L^{(j)} \cup D_A^{(j)} \quad \text{and} \quad D_U^{(j+1)} = D_U^{(j)} \setminus D_A^{(j)}$ 
    \ENDIF
    \ENDFOR
  \end{algorithmic}
  \end{spacing}
\end{algorithm}

\minisection{Reduction to standard DAL and IMP.} \method subsumes both of its parent protocols, DAL and LTH, as
special cases of \Cref{alg:method}, which clarifies the structural correspondence between the two loops and
pinpoints where the computational savings originate. In either reduction, the mask reactivation (line~5) and
the finetuning phase (lines~9) are dropped, as neither operation is part of the respective baseline. A \emph{standard DAL loop} is recovered by fixing
$M^{(j)}_{\text{init}} = \mathbf{1}$ throughout and removing the pruning step (line~8) together with the
rewind selection (lines~11--13), so that every iteration trains a dense model from the same initialization
and passes it directly to the acquisition function. Conversely, a \emph{standard IMP procedure} is obtained
by freezing the training data at $D_L^{(j)} = D_L^{(0)}$ for all $j$ and removing the acquisition update
(lines~15--18): the loop then reduces to repeated train-and-prune rounds on a fixed dataset, exactly matching
the original LTH formulation of~\cite{frankle2018the}.

\subsection{ResNet-18 Backbone Adaptations}\label{app:resnet_adaptations}
The standard ResNet-18 stem is designed for $224{\times}224$ input resolutions and
down-samples the spatial resolution before the first
residual block. Applying this stem unchanged to substantially smaller
inputs would collapse the feature maps prematurely. We therefore
reduce the down-sampling as~\citep{praka} in the stem for lower resolutions while
keeping all residual blocks and the final $512$-dimensional embedding
identical. \Cref{tab:resnet_stem} lists the per-resolution
configurations; convolution padding is always set to
$\lfloor K/2 \rfloor$.

\begin{table}[H]
    \centering
    \caption{ResNet-18 stem adaptations dependent on the targeted input resolution.
    All variants share identical network topology and penultimate feature dimensionality (512); 
    differences are restricted to the input stem and initial pooling}\label{tab:resnet_stem}
    \begin{tabular}{l@{\hskip 0.5in}c@{\hskip 0.7in}c@{\hskip 0.5in}c}
        \toprule
        & \textbf{CIFAR} & \textbf{Tiny-ImageNet} & \textbf{ImageNet} \\
        Input Scale & $32{\times}32$ & $64{\times}64$ & $224{\times}224$ \\
        \midrule
        Conv kernel size  & $3{\times}3$ & $7{\times}7$ & $7{\times}7$ \\
        Conv stride       & $1$          & $2$          & $2$          \\
        Conv padding      & $1$          & $3$          & $3$          \\
        Max-pool          & ---          & ---          & $3{\times}3$, stride $2$, pad $1$ \\
        \midrule
        Feature map after stem & $32{\times}32$ & $32{\times}32$ & $56{\times}56$ \\
        \bottomrule
    \end{tabular}
\end{table}

\subsection{Winning-Ticket Significance Testing}\label{app:wilcoxon}
We assess whether a pruned method qualifies as a winning ticket with respect to the Dense reference. Tests are performed for each acquisition function and dataset separately with a one-sided paired Wilcoxon test, no power corrections are applied. $H_1$ states that the pruned variant is non-inferior to Dense up to a tolerance of $\epsilon=-0.005$ AULC:
\begin{equation}
  H_1:\ \mathrm{AULC}_{\text{Dense}} - \mathrm{AULC}_{\text{method}} < \epsilon \,,  
\end{equation}
where $\mathrm{AULC}$ denotes the area under the accuracy-vs-labels learning curve across the DAL iterations of a single seeded run. The AULC summarizes the entire learning trajectory in a single scalar and is an aggregate for active learning benchmarks~\citep{werner2024a}. Concretely, let $n_0 < n_1 < \dots < n_A$ denote the cumulative number of labeled samples after each of the $A{+}1$ DAL iterations (including the initial pool) and let $a_i$ be the corresponding test accuracy. The AULC is the trapezoidal integral of the accuracy--labels curve, normalized by the label
range:
\begin{equation}\label{eq:aulc}
    \mathrm{AULC}
    = \frac{1}{n_A - n_0}
      \sum_{i=1}^{A}
      \frac{a_{i-1} + a_i}{2}\,
      \bigl(n_i - n_{i-1}\bigr)\,.
\end{equation}
Averaging over DAL rounds rather than reporting the final-iteration accuracy alone makes the test sensitive to trajectory-level differences, not just to the endpoint.

We employ a one-sided seed-paired test on the $\mathrm{AULC}_{\text{Dense}} - \mathrm{AULC}_{\text{method}}$ differences. Paired seeds imply identical initialization and initial data splits. The Wilcoxon test makes no assumption on the distribution of the AULC differences. Rejection at $p < 0.05$ means that significant underperformance of the method relative to the dense baseline cannot be established at the seed-level variability; consequently the method is then marked as a \emph{winning ticket}. We omit values for Places365 as $3$ seeds provide insufficient power for the statistical test. The non-inferiority margin $\epsilon$ is a deliberate design choice with no standard value, and the verdict can be sensitive to it: in the CIFAR-100 HB regime, the few cases that do not pass at the strict $\epsilon=0.005$ exhibit a mean paired AULC gap to Dense of ${\approx}\,0.007$, and relaxing the tolerance to the still-small $\epsilon=0.01$ recovers non-inferiority. We report the strict setting throughout to remain conservative.

\subsection{Hyperparameter Sensitivity}\label{app:hyperparam_search}
We probe the three key hyperparameters of \method, e.g. pruning rate $p_c$, reactivation rate $p_r$ and target sparsity $s$ on CIFAR-100  with margin sampling with 3 seeds: the per-round pruning rate $p_{\text{c}}$ (\Cref{tab:hyperparam_prune_rate}) and the target sparsity, expressed as the retained active-parameter fraction $1-s$ (\Cref{tab:hyperparam_target_sparsity}) together with the reactivation rate jointly control how the resulting sparsity evolves across the DAL sequence. Across both sweeps, \method showed stable performance with limited sensitivity to $p_c$ and $p_r$. Only under extreme pruning conditions with roughly $99.95\%$ of all original parameters removed, the accuracy of the model starts to collapse.

\begin{table}[h]
    \centering
    \caption{\emph{Per-round pruning-rate sweep.} Final classification accuracy $A_9$ on CIFAR-100 (Margin, 3 seeds) as a function of the per-round pruning rate $p_{\text{c}}$.}
    \label{tab:hyperparam_prune_rate}
    \begin{tabular}{cc ccccc}
        \toprule
        \textbf{Budget} & \textbf{Method} & \multicolumn{5}{c}{\textbf{Pruning rate} $p_{\text{c}}$} \\
        \cmidrule(lr){3-7}
        & & $0.50$ & $0.66$ & $0.75$ & $0.85$ & $0.95$ \\
        \midrule
        \multirow{4}{*}{LB} 
            & Dense          & \multicolumn{5}{c}{$\mathbf{18.66 {\scriptstyle \pm 1.68}}$} \\ \cmidrule(lr){2-7}
            & Rand Sparse    & $15.34 {\scriptstyle \pm 0.89}$ & $13.02 {\scriptstyle \pm 1.28}$ & $13.44 {\scriptstyle \pm 0.90}$ & $12.66 {\scriptstyle \pm 0.52}$ & $13.32 {\scriptstyle \pm 0.48}$ \\
            & Rand Prune     & $16.03 {\scriptstyle \pm 0.78}$ & $11.59 {\scriptstyle \pm 1.56}$ & $12.11 {\scriptstyle \pm 1.26}$ & $12.64 {\scriptstyle \pm 0.17}$ & $13.16 {\scriptstyle \pm 0.72}$ \\
            & \method (ours) & $\mathbf{19.15 {\scriptstyle \pm 1.27}}$ & $\mathbf{18.66 {\scriptstyle \pm 0.70}}$ & $\mathbf{18.93 {\scriptstyle \pm 1.48}}$ & $\mathbf{18.94 {\scriptstyle \pm 0.84}}$ & $\mathbf{19.22 {\scriptstyle \pm 0.20}}$ \\
        \midrule
        \multirow{4}{*}{HB}
            & Dense          & \multicolumn{5}{c}{$\mathbf{53.27 {\scriptstyle \pm 1.13}}$}  \\ \cmidrule(lr){2-7}
            & Rand Sparse    & $46.26 {\scriptstyle \pm 0.50}$ & $42.16 {\scriptstyle \pm 0.47}$ & $41.66 {\scriptstyle \pm 0.34}$ & $41.63 {\scriptstyle \pm 1.39}$ & $42.10 {\scriptstyle \pm 0.49}$ \\
            & Rand Prune     & $27.81 {\scriptstyle \pm 3.40}$ & $17.04 {\scriptstyle \pm 1.71}$ & $17.00 {\scriptstyle \pm 0.34}$ & $17.42 {\scriptstyle \pm 1.38}$ & $17.44 {\scriptstyle \pm 1.36}$ \\
            & \method (ours) & $\mathbf{53.13 {\scriptstyle \pm 0.86}}$ & $\mathbf{52.17 {\scriptstyle \pm 1.01}}$ & $\mathbf{52.60 {\scriptstyle \pm 0.83}}$ & $\mathbf{52.05 {\scriptstyle \pm 0.97}}$ & $\mathbf{52.31 {\scriptstyle \pm 0.59}}$ \\
        \bottomrule
    \end{tabular}
\end{table}

\begin{table}[h]
    \centering
    \caption{\emph{Target-sparsity sweep.} Final classification accuracy $A_9$ on CIFAR-100 (Margin, 3 seeds) as a function of the retained active-parameter fraction $1-s$ and the coupled reactivation budget $p_{\text{r}}$; the per-round pruning rate is held fixed at $p_{\text{c}}=0.66$. Bold marks \method runs that match or exceed Dense within one standard deviation.}
    \label{tab:hyperparam_target_sparsity}
    \begin{tabular}{ccccc}
        \toprule
        \textbf{Method} & \textbf{Active fraction} & \textbf{Reactivation} & \multicolumn{2}{c}{\textbf{$\mathbf{A_9}$ [\%]}} \\
        \cmidrule(lr){4-5}
        & $1-s$ & $p_{\text{r}}$ & LB & HB \\
        \midrule
        Dense          & \xmark       & \xmark      & $\mathbf{18.66 {\scriptstyle \pm 1.68}}$ & $\mathbf{53.27 {\scriptstyle \pm 1.13}}$ \\
        \midrule
        \multirow{11}{*}{\method (ours)} & $0.0005$ & $0.001$ & $14.94 {\scriptstyle \pm 0.25}$          & $27.59 {\scriptstyle \pm 18.37}$ \\
         & $0.001$  & $0.002$ & $\mathbf{17.73 {\scriptstyle \pm 0.51}}$ & $45.77 {\scriptstyle \pm 0.24}$ \\
         & $0.0025$ & $0.005$ & $\mathbf{19.23 {\scriptstyle \pm 0.55}}$ & $48.96 {\scriptstyle \pm 0.18}$ \\
         & $0.005$  & $0.010$ & $\mathbf{18.77 {\scriptstyle \pm 0.77}}$ & $48.97 {\scriptstyle \pm 1.08}$ \\
         & $0.01$   & $0.020$ & $\mathbf{18.54 {\scriptstyle \pm 0.66}}$ & $49.42 {\scriptstyle \pm 1.11}$ \\
         & $0.025$  & $0.050$ & $\mathbf{19.01 {\scriptstyle \pm 0.83}}$ & $51.18 {\scriptstyle \pm 1.59}$ \\
         & $0.05$   & $0.100$ & $\mathbf{18.66 {\scriptstyle \pm 0.70}}$ & $\mathbf{52.17 {\scriptstyle \pm 1.01}}$ \\
         & $0.07$   & $0.100$ & $\mathbf{18.66 {\scriptstyle \pm 0.70}}$ & $\mathbf{52.17 {\scriptstyle \pm 1.01}}$ \\
         & $0.10$   & $0.100$ & $\mathbf{19.17 {\scriptstyle \pm 1.22}}$ & $\mathbf{52.52 {\scriptstyle \pm 1.52}}$ \\
         & $0.15$   & $0.100$ & $\mathbf{18.68 {\scriptstyle \pm 1.04}}$ & $\mathbf{52.68 {\scriptstyle \pm 0.81}}$ \\
         & $0.20$   & $0.100$ & $\mathbf{19.00 {\scriptstyle \pm 1.39}}$ & $\mathbf{53.04 {\scriptstyle \pm 0.31}}$ \\
        \bottomrule
    \end{tabular}
\end{table}

\subsection{Comparison to Pruning-at-Initialization and Dynamic Sparse Training}\label{app:pai_rigl}
\method derives its mask from a trained dense model and refines it across DAL iterations. Two other competing
paradigms avoid dense train-and-prune cycles altogether. \emph{Pruning at initialization} (PAI) fixes a
sparse mask \emph{before} any training. Three common pruning strategies used for the fixed mask are: connection sensitivity
(SNIP,~\citealp{lee2019snip}), gradient-flow preservation (GraSP,~\citealp{wang2020picking}), or
iterative synaptic-flow conservation (SynFlow,~\citealp{tanaka2020pruning}). Further, \emph{dynamic sparse training} (RigL,~\citealp{evci2020rigging}) maintains a fixed sparsity budget throughout training while periodically
regrowing and pruning connections based on gradient magnitude. Both families achieve the target sparsity $s$ without a dense
training phase within a single AL iteration.

\Cref{tab:pai_rigl} compares these methods against \method under the Margin acquisition function on
CIFAR-100, in both the low-budget (LB) and high-budget (HB) regimes. All sparse methods are matched to the same
target sparsity $s$ and evaluated within the identical DAL loop; a major difference is \textbf{how} and \textbf{when} the pruning mask is obtained. PAI masks are recomputed from scratch at the \textbf{start} of each AL iteration, whereas RigL computes it \textbf{during} network training and \method performs the pruning \textbf{after} initial iteration training and before finetuning.

\begin{table}[h]
    \centering
    \caption{\emph{Comparison to pruning-at-initialization and dynamic sparse training.} Final classification
    accuracy [\%] on CIFAR-100 with the Margin acquisition function, for the low-budget (LB) and high-budget (HB)
    regimes. All sparse methods are pruned to the same target sparsity $s=0.95\%$. Prune paradigm column
    defines at which point the pruning is performed. Pruning at Initialization (PAI), while joint optimizes the pruning mask
    and the network training simultaneously.}
    \label{tab:pai_rigl}
    \begin{tabular}{llcc}
        \toprule
        \textbf{Method} & \textbf{Prune} & \textbf{Margin LB} & \textbf{Margin HB} \\
        \midrule 
        Dense          & \xmark        & $17.45 {\scriptstyle \pm 2.07}$ & $52.48 {\scriptstyle \pm 1.33}$ \\
        \midrule
        SNIP           & PAI  & $16.82 {\scriptstyle \pm 2.36}$ & $49.91 {\scriptstyle \pm 1.04}$ \\
        SynFlow        & PAI  & $16.31 {\scriptstyle \pm 1.95}$ & $50.54 {\scriptstyle \pm 1.67}$ \\
        GraSP          & PAI  & $13.02 {\scriptstyle \pm 1.04}$ & $48.39 {\scriptstyle \pm 1.27}$ \\
        \midrule
        RigL           & while training  & $14.52 {\scriptstyle \pm 0.81}$ & $54.25 {\scriptstyle \pm 1.07}$ \\
        \midrule
        One-Shot & posthoc        & $17.52 {\scriptstyle \pm 2.51}$ & $52.18 {\scriptstyle \pm 1.14}$ \\
        \method (ours) & iter. posthoc        & $17.52 {\scriptstyle \pm 1.80}$ & $52.29 {\scriptstyle \pm 1.30}$ \\
        \bottomrule
    \end{tabular}
    \vspace{-1em}
\end{table}

\minisection{RigL.} Dynamic sparse training surpasses Dense in the high-budget regime, where enough labeled data is available to jointly fit the mask topology and the parameters. This benefit carries two practical costs. First, the periodic regrow step requires the gradient over the \emph{inactive} weights, so every update reverts to a dense backward pass even though the forward path is sparse. Second, the mask keeps shifting throughout training, which obstructs the structured patterns -- in particular the N:M masks of~\citet{mishra2021accelerating} -- that translate unstructured sparsity into measurable hardware speedups; \method, whose mask stabilizes within a DAL iteration, is comparatively well-aligned with such patterns. The ranking inverts at low budget. With few labels, the externally imposed magnitude mask of \method appears to act as a regularizer, while RigL's joint mask--weight optimization is more prone to overfit the small labeled pool.

\minisection{SNIP, SynFlow, GraSP.} The three pruning at initialization criteria show moderate
underperformance relative to Dense and \method at both budgets while remaining clearly above the random
sparsity baselines reported in \Cref{tab:final_acc_results}. This matches the broader evidence
collected by~\citet{hoefler2021sparsity} that parameter importance estimated \emph{before} training is a
weak proxy for importance after convergence: the saliency landscape shifts as the network fits the data,
so a mask fixed at initialization cannot capture the structure that only emerges from training.

\minisection{One-Shot Pruning.} Post-hoc one-shot pruning reaches accuracy parity with Dense and
\method at both budgets, indicating that magnitude-based pruning is competitive once the model has been
trained. Because the mask is derived only after training, however, every DAL iteration trains the network
in its dense form and cannot exploit the sparse-kernel savings that \method realizes from the second AL
round onward. Iterative pruning has also been reported to outperform one-shot pruning at very high
sparsities~\citep{wojnar2024how}, suggesting that the parity observed at our target sparsity (95\%) may not
extend to higher sparsities.

\subsection{Second-Stage Training Cost and Recovery}\label{app:stage2_cost}

\begin{table}[t]
    \centering

    \caption{Finetuning training cost across all experimental settings.
    Training (T) and Finetuning (F) columns report the mean ($\pm$ std) number of training
    epochs per DAL iteration, averaged across acquisition functions, DAL rounds,
    and seeds. \emph{Conv.}\ gives the fraction of iterations that re-reached
    the training-accuracy convergence threshold before hitting the patience
    budget; values below $100\%$ indicate runs that were terminated at the
    patience ceiling.}
    \label{tab:stage2_cost_full}
    
    \begin{tabular}{lllccc}
        \toprule
        \textbf{DAL Setting} & \textbf{Patience} & \textbf{Train}
            & \textbf{T. epochs} & \textbf{F. epochs}
            & \textbf{Conv. (\%)} \\
        \midrule
        \multirow{3}{*}{CIFAR-100 LB} & \multirow{3}{*}{$\varphi=30$}
            & Dense       & $51.1 {\scriptstyle\pm 18.8}$ & $3.0 {\scriptstyle\pm 3.5}$   & $100.0$ \\
            &  & I\&P (ours) & $43.1 {\scriptstyle\pm 16.0}$ & $2.6 {\scriptstyle\pm 3.1}$   & $100.0$ \\
            &  & Rand Prune  & $80.0 {\scriptstyle\pm 28.0}$ & $29.6 {\scriptstyle\pm 4.4}$  & $11.0$  \\
        \midrule
        \multirow{3}{*}{CIFAR-100 HB}  & \multirow{3}{*}{$\varphi=30$}
            & Dense       & $80.1 {\scriptstyle\pm 16.3}$ & $1.8 {\scriptstyle\pm 1.7}$   & $100.0$ \\
            & & I\&P (ours) & $82.6 {\scriptstyle\pm 18.8}$ & $1.9 {\scriptstyle\pm 1.0}$   & $100.0$ \\
            & & Rand Prune  & $124.3 {\scriptstyle\pm 36.1}$ & $30.5 {\scriptstyle\pm 1.7}$ & $8.0$  \\
        \midrule
        \multirow{3}{*}{Imagewoof LB}  & \multirow{3}{*}{$\varphi=60$}
            & Dense       & $82.7 {\scriptstyle\pm 22.5}$ & $7.8 {\scriptstyle\pm 7.9}$   & $100.0$ \\
            & & I\&P (ours) & $61.5 {\scriptstyle\pm 21.3}$  & $4.2 {\scriptstyle\pm 3.1}$   & $100.0$ \\
            & & Rand Prune  & $104.4 {\scriptstyle\pm 30.0}$ & $56.4 {\scriptstyle\pm 10.0}$ & $22.7$  \\
        \midrule
        \multirow{3}{*}{Imagewoof HB}  & \multirow{3}{*}{$\varphi=60$}
            & Dense       & $103.8 {\scriptstyle\pm 20.3}$ & $5.2 {\scriptstyle\pm 5.3}$   & $100.0$ \\
            & & I\&P (ours) & $95.7 {\scriptstyle\pm 21.4}$  & $5.8 {\scriptstyle\pm 4.1}$  & $100.0$ \\
            & & Rand Prune  & $132.6 {\scriptstyle\pm 32.7}$ & $57.6 {\scriptstyle\pm 8.5}$ & $17.5$  \\
        \midrule
        \multirow{3}{*}{Tiny-ImageNet}  & \multirow{3}{*}{$\varphi=30$}
            & Dense       & $63.1 {\scriptstyle\pm 16.6}$ & $3.3 {\scriptstyle\pm 4.3}$  & $99.9$  \\
            & & I\&P (ours) & $56.5 {\scriptstyle\pm 19.6}$ & $3.8 {\scriptstyle\pm 2.3}$  & $100.0$ \\
            & & Rand Prune  & $80.7 {\scriptstyle\pm 19.6}$ & $29.8 {\scriptstyle\pm 4.0}$ & $9.5$   \\
        \midrule
        \multirow{2}{*}{\shortstack[l]{Places365\\ResNet-50 (sup.)}}  & \multirow{2}{*}{$\varphi=15$}
            & Dense       & $35.6 {\scriptstyle\pm 6.1}$ & $1.0 {\scriptstyle\pm 0.2}$ & $100.0$ \\
            & & I\&P (ours) & $34.4 {\scriptstyle\pm 5.8}$ & $1.0 {\scriptstyle\pm 0.0}$ & $100.0$ \\
        \midrule
        \multirow{2}{*}{\shortstack[l]{Places365\\ConvNext v2 Tiny (ssl.)}}  & \multirow{2}{*}{$\varphi=15$}
            & Dense       & $28.2 {\scriptstyle\pm 6.2}$ & $1.0 {\scriptstyle\pm 0.0}$ & $100.0$ \\
            & & I\&P (ours) & $28.5 {\scriptstyle\pm 5.5}$ & $1.6 {\scriptstyle\pm 0.8}$ & $100.0$ \\
        \midrule
        \multirow{2}{*}{\shortstack[l]{Places365\\DeiT-Small (sup.)}} & \multirow{2}{*}{$\varphi=15$}
            & Dense       & $22.4 {\scriptstyle\pm 9.7}$ & $1.4 {\scriptstyle\pm 1.4}$ & $100.0$ \\
            & & I\&P (ours) & $23.9 {\scriptstyle\pm 9.5}$ & $2.4 {\scriptstyle\pm 1.1}$ & $100.0$ \\
        \midrule
        \multirow{2}{*}{\shortstack[l]{Places365\\DINO ViT-S (ssl.)}} & \multirow{2}{*}{$\varphi=15$}
            & Dense       & $41.1 {\scriptstyle\pm 9.8}$ & $4.9 {\scriptstyle\pm 6.2}$ & $76.7$ \\
            & & I\&P (ours) & $41.7 {\scriptstyle\pm 9.1}$ & $3.1 {\scriptstyle\pm 5.1}$ & $86.7$ \\
        \bottomrule
    \end{tabular}
\end{table}

\Cref{tab:stage2_cost_full} quantifies the additional training cost introduced by the finetuning phase. The \emph{convergence rate} records whether the pruned model re-reached the training-accuracy threshold within the patience $\varphi$, making it a direct indicator of whether the pruned model remains trainable to its original performance level. Across all settings, \method re-converges quickly (100\% on CIFAR-100, Imagewoof, Tiny-ImageNet, and Places365 except DINO), using a marginal fraction of the main training budget. The overhead remains well below $< 10\%$ in almost all settings and is indistinguishable from the Dense dual-stage baseline, that skips the pruning step and terminates on the next epoch that is again above the early stopping criterion.

In sharp contrast, Rand Prune exhibits the opposite pattern: its finetuning phase rarely re-converges within the patience $\varphi$ ceiling in the vast majority of runs. High variance in finetuning epoch is predominantly determined by non converging trainings. Read together, the two observations tell a single story: finetuning is inexpensive. It is cheap for \method precisely because the magnitude-based pruning preserves the performance of the resulting subnetwork while random decision fail to.

The only outlier in the table is DINO ViT-S on Places365, where both the dense baseline ($76.7\%$) and I\&P ($86.7\%$) occasionally fail to converge within the patience budget. This reflects a property of an optimization problem of the network and an inadequate early stopping criteria.

In \Cref{fig:finetune_discounted}, the triangle markers depict Dense and \method with their
finetuning phase accounted as free (finetune \mbox{$\text{MACs}=0$}). \Cref{tab:finetune_ablation}
reports the corresponding final accuracy when the finetune phase is omitted altogether: Dense is
essentially unaffected, whereas \method loses ${\approx}\,1\%$ absolute accuracy, indicating that the
finetune phase is relevant for recovering the function drift induced by pruning.

\begin{minipage}{\textwidth}
  \begin{minipage}[b]{0.49\textwidth}
    \begin{figure}[H]
    \centering
    \includegraphics[width=1.0\linewidth]{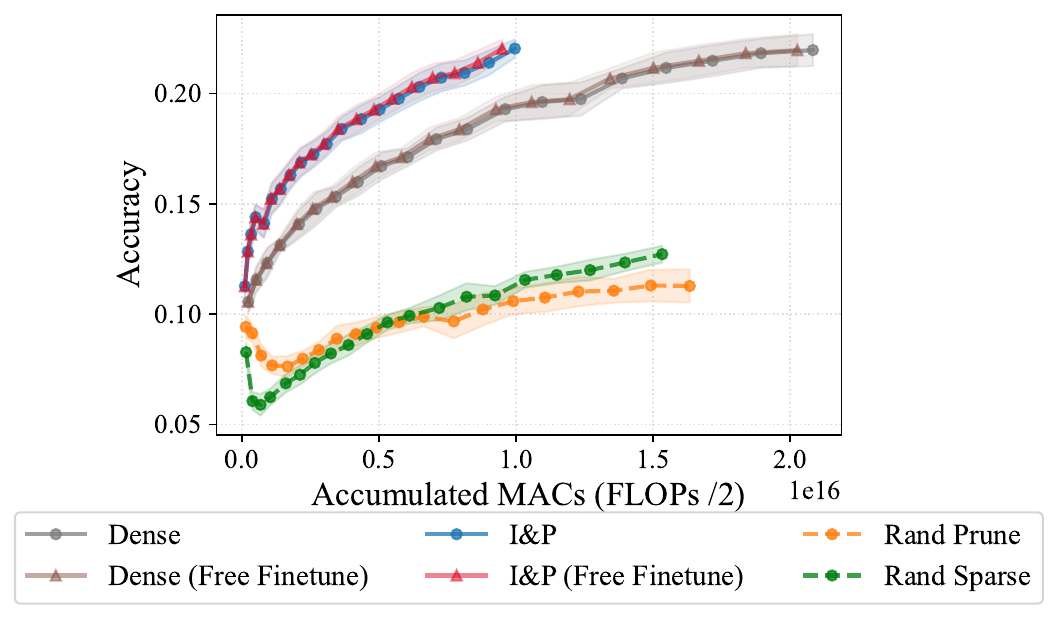}
    \caption{\emph{Mean Accuracy with one std as a function of cumulative MACs} calculated over \num{15} seeds with $\mathcal{S}$ = Margin  on Tiny-ImageNet. Triangle markers do not account for the needed finetune MACs.}
    \label{fig:finetune_discounted}
    \end{figure}
  \end{minipage}
  \hfill
  \begin{minipage}[b]{0.49\textwidth}
    \centering
    \captionof{table}{\emph{Finetuning ablation.} Final classification accuracy $A_9$ on CIFAR-100
    (Margin, 3 seeds) with and without the per-round finetuning phase. Dense is essentially unaffected,
    whereas removing finetuning from \method drops by ${\approx}\,1\%$.}
    \begin{tabular}{lcc}
      \toprule
      \textbf{Method} & \textbf{$A_9$ with FT} & \textbf{$A_9$ without FT} \\
      \midrule
      Dense           & ${53.27 {\scriptstyle\pm 1.13}}$ & ${53.20 {\scriptstyle\pm 0.76}}$ \\
      Rand Sparse     & ${42.16 {\scriptstyle\pm 0.46}}$ & ${41.11 {\scriptstyle\pm 3.24}}$ \\
      Rand Prune      & ${17.04 {\scriptstyle\pm 1.71}}$ & ${9.93 {\scriptstyle\pm 0.01}}$ \\
      \method (ours)  & ${52.17 {\scriptstyle\pm 1.01}}$ & ${51.20 {\scriptstyle\pm 0.76}}$ \\
      \bottomrule
    \end{tabular}
    \vspace{4.5em}
    \label{tab:finetune_ablation}
    \end{minipage}
  \end{minipage}

\newpage
\subsection{Mask Trajectories across Acquisition Functions}\label{app:mask_trajectory}
To compare the sparse subnetworks discovered under different acquisition functions, we quantify the overlap of their final pruning masks with the \emph{Jaccard similarity} $J$. Let $M_a, M_b \in \{0,1\}^d$ denote two binary masks over the flattened $d$ prunable parameters, where $M_{\cdot,i}=1$ marks an active (retained) weight. The Jaccard similarity is the size of the intersection of the active supports over the size of their union,

\begin{equation}\label{eq:jaccard}
    J(M_a, M_b)
    = \frac{\lvert M_a \cap M_b \rvert}{\lvert M_a \cup M_b \rvert}
    = \frac{\sum_{i} \mathbf{1}\!\left[M_{a,i}=1 \,\wedge\, M_{b,i}=1\right]}
           {\sum_{i} \mathbf{1}\!\left[M_{a,i}=1 \,\vee\, M_{b,i}=1\right]}\,,
\end{equation}

with $J=1$ for identical masks and $J=0$ for disjoint ones. Unlike a hamming distance, the Jaccard measure ignores the typically large set of pruned weights and thus isolates the agreement among the weights that remain active.

\vspace{-0.75em}
\begin{minipage}{\textwidth}
  \begin{minipage}[b]{0.45\textwidth}
    \begin{figure}[H]
    \centering
    \includegraphics[width=1.0\linewidth]{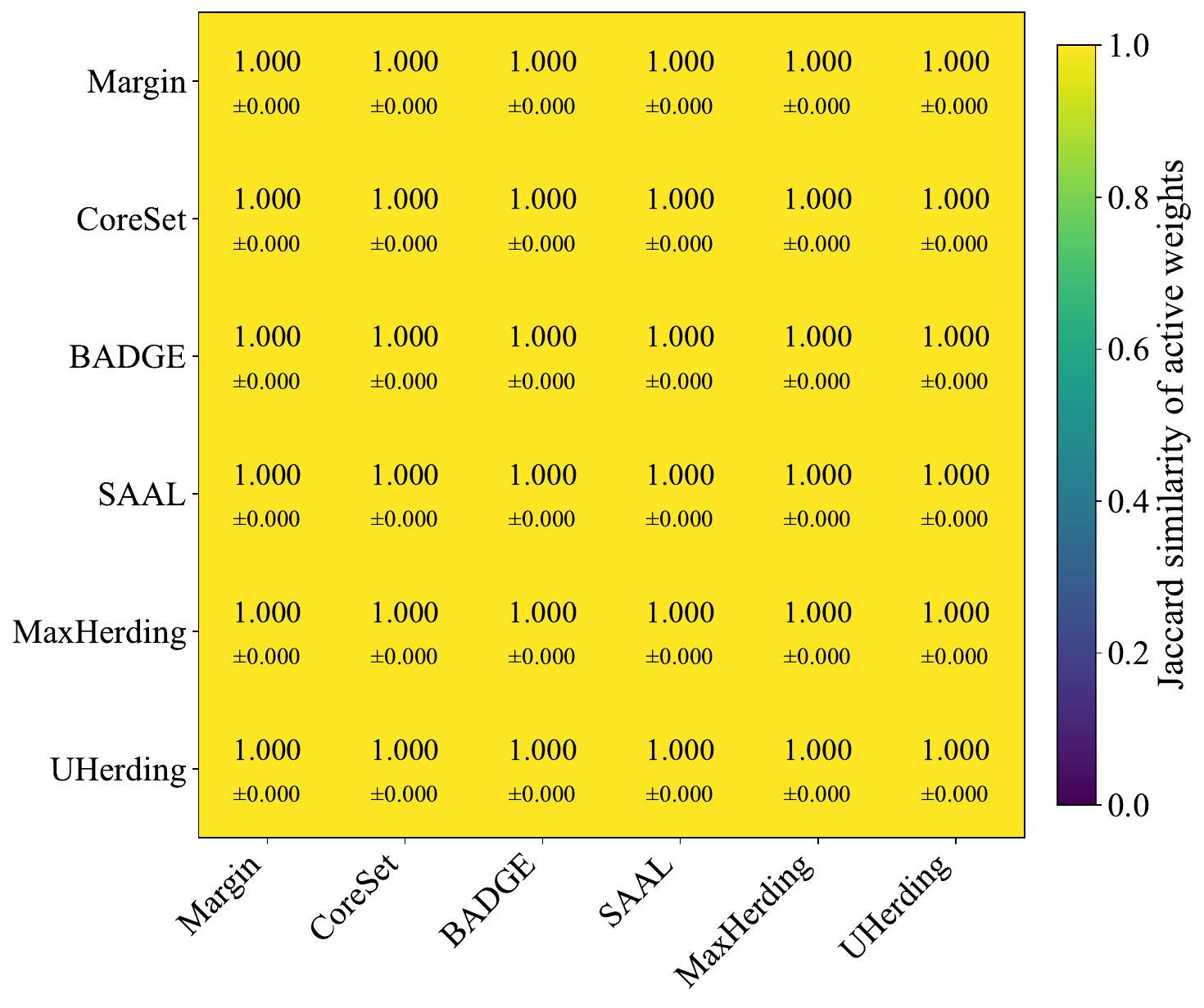}
    \caption{After AL Iteration $j=0$}
    \label{fig:al_mask_0}
    \end{figure}
    \vspace{-0.75em}
    \begin{figure}[H]
    \centering
    \includegraphics[width=1.0\linewidth]{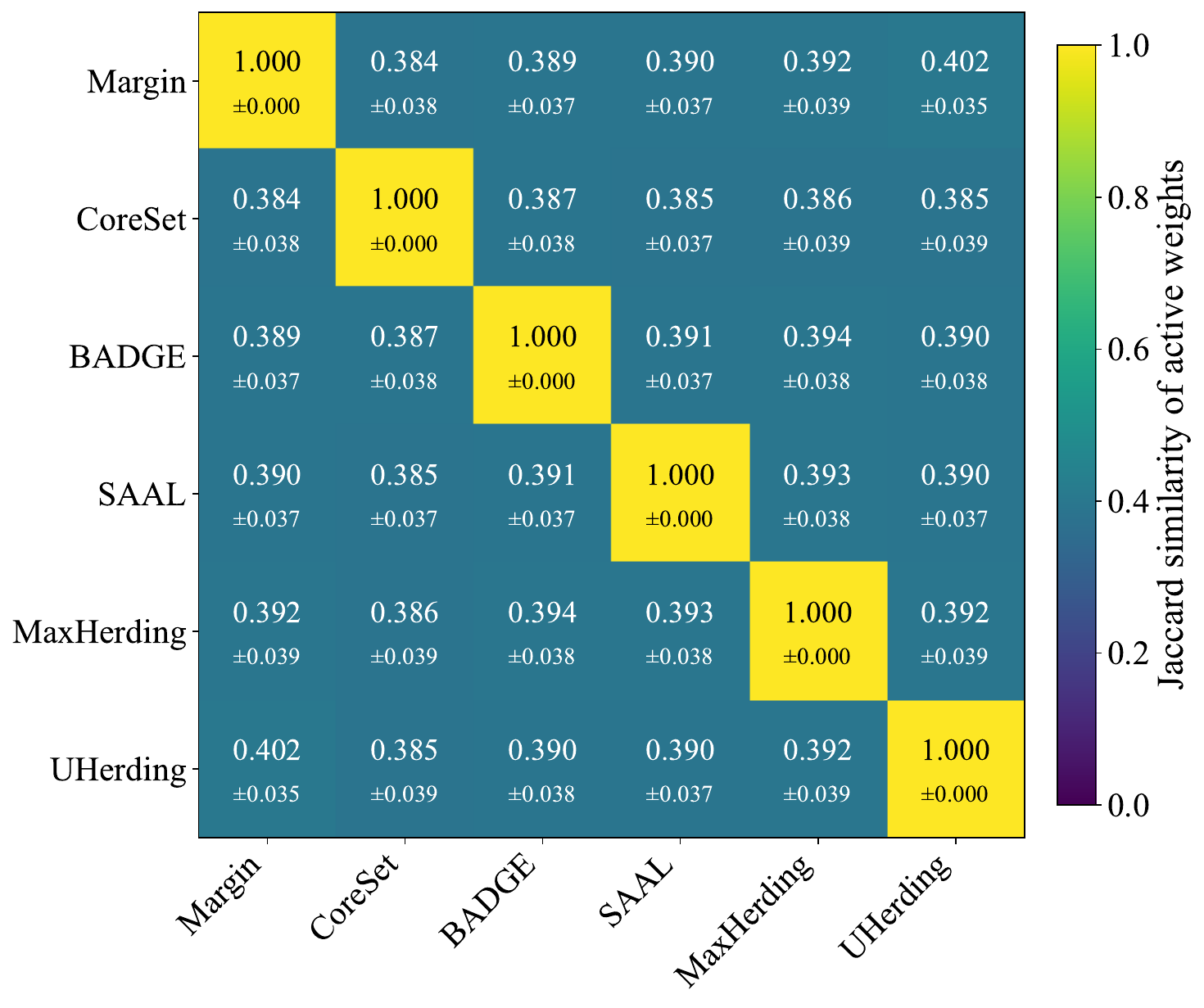}
    \caption{After AL Iteration $j=2$}
    \label{fig:al_mask_2}
    \end{figure}
    \vspace{-0.75em}
  \end{minipage}
  \hfill
  \begin{minipage}[b]{0.45\textwidth}
    \begin{figure}[H]
    \centering
    \includegraphics[width=1.0\linewidth]{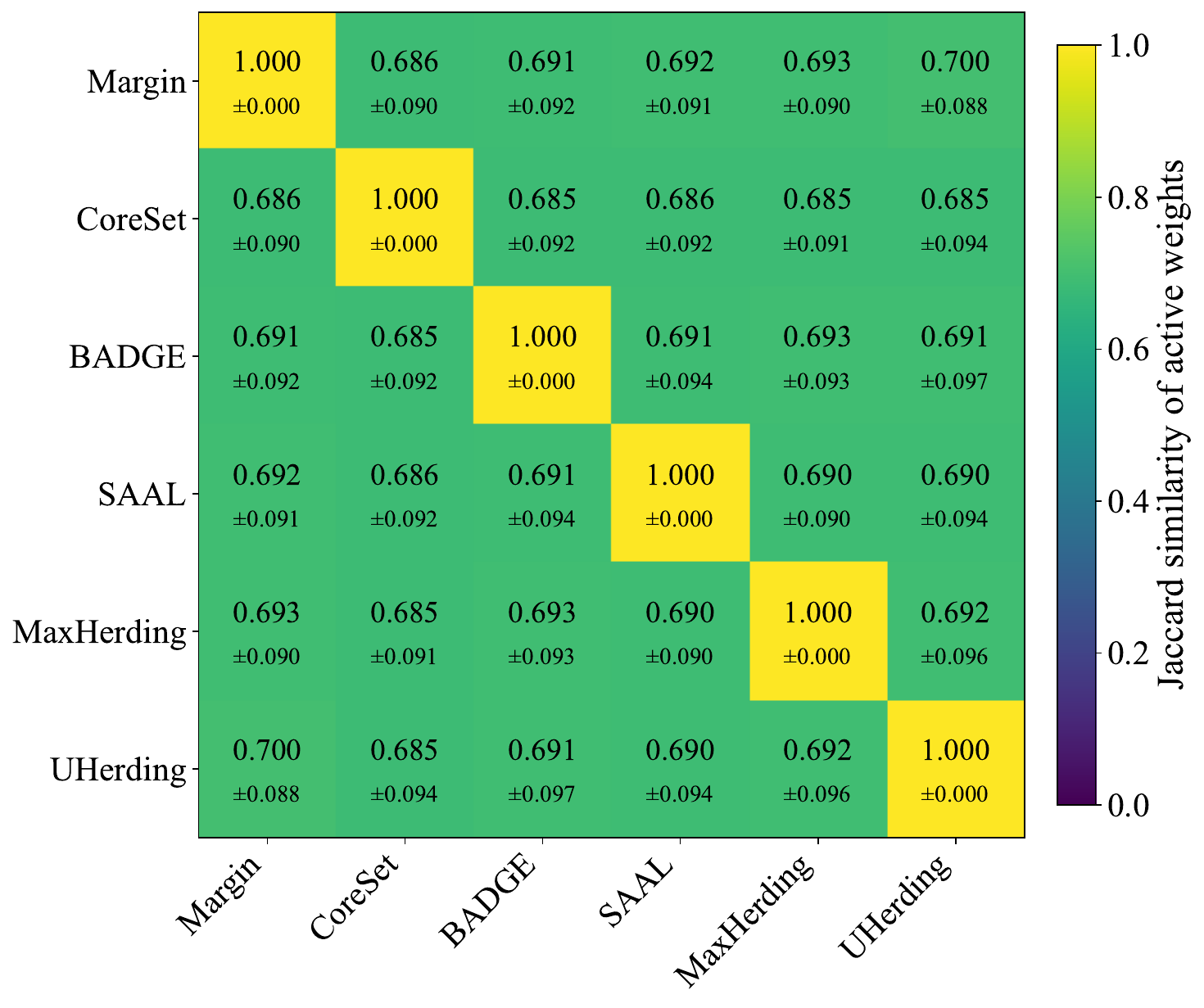}
    \caption{After AL Iteration $j=1$}
    \label{fig:al_mask_1}
    \end{figure}
    \vspace{-0.75em}
    \begin{figure}[H]
    \centering
    \includegraphics[width=1.0\linewidth]{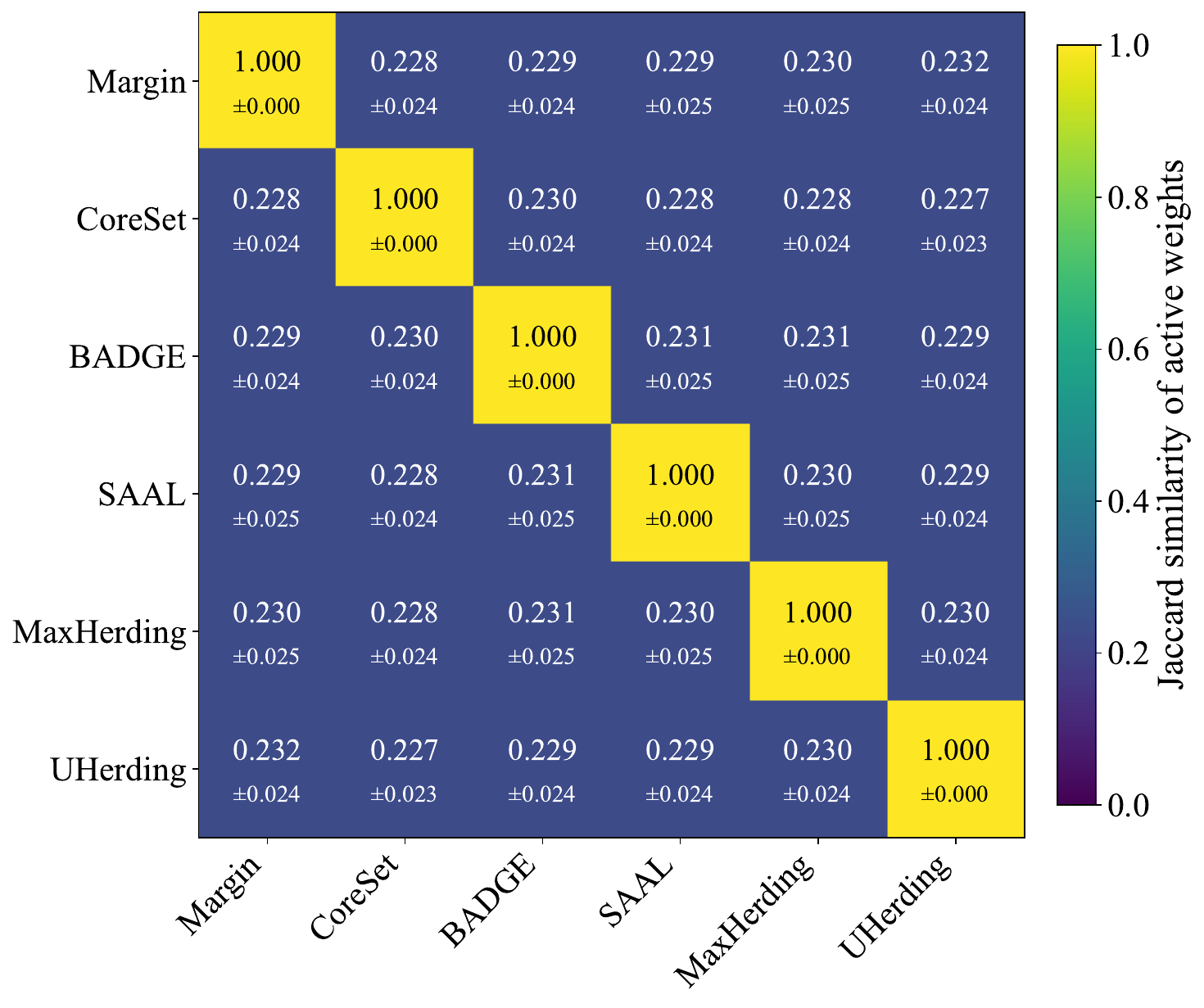}
    \caption{After AL Iteration $j=9$}
    \label{fig:al_mask_3}
    \end{figure}
    \vspace{-0.75em}
  \end{minipage}
\end{minipage}

\Cref{fig:al_mask_0,fig:al_mask_1,fig:al_mask_2,fig:al_mask_3} illustrate the pairwise Jaccard similarity between the pruning masks $M^{(j+1)}$ across DAL iterations $j$. Results are shown for all six acquisition functions and averaged over 10 seeds in the CIFAR-HB setting. Similarities are only computed between matching seeds from identical initialization. At iteration~$j=0$ all masks coincide exactly ($J=1$), because training and pruning are performed on the identical initial pool $D_L^{(0)}$ and the acquisition function only enters at the end of the iteration. From iteration~$j=1$ onward the masks diverge sharply, yet the inter-function similarity is still well above chance. At a sparsity of $s=0.95$ two independent random masks would result in an expected Jaccard similarity of $(1-s)/(1+s) \approx 0.0256$, whereas the observed final similarity sits roughly an order of magnitude higher. We interpret this as evidence that the six acquisition strategies converge on a shared coreset of parameters and fit a comparatively large, acquisition-specific number of parameters around it that are not essential for the task. This might be an indication that the target sparsity could be raised without major adverse effects (see Appendix~\ref{app:hyperparam_search} for higher sparsities in same the scenario).

\subsection{Sparse-Aware FLOPs Estimation}\label{app:flops_details}
This section provides the full per-layer formulas underlying the mask-aware MAC (e.g FLOPs / 2) estimation. Pruning is implemented via \texttt{torch.nn.utils.prune} module, which attaches binary masks to parameters without physically removing weights. All sparsities are read directly from these masks.

For each layer $\ell$, we compute the dense MACs from the layer geometry
(kernel size, channel counts, and output spatial dimensions) and scale by the
weight density $d_w^{(\ell)} = 1 - s_w^{(\ell)}$ obtained from the pruning
mask to obtain the sparse count. Bias additions are scaled analogously by the
per-layer bias density $d_b^{(\ell)}$. Operations not attributable to
parametric layer forward hooks, notably attention matmuls ($QK^\top$ and
$\text{attn}\cdot V$), are captured via a global MAC counter and included at
full cost in both the dense and sparse totals. The total sparse MAC count is
therefore
\begin{equation}\label{eq:sparse_flops}
    C_{\text{sparse}}
    = C_{\text{global}}
    - C_{\text{hook}}^{\text{dense}}
    + C_{\text{hook}}^{\text{sparse}}\,,
\end{equation}

\minisection{Convolutional layers.}
For a \texttt{Conv2d} layer $\ell$ with kernel size $K_h \times K_w$,
$C_{\text{in}}$ input channels, $C_{\text{out}}$ output channels, $G$ groups,
and output spatial dimensions $H_{\text{out}} \times W_{\text{out}}$, the dense
weight MACs are
\begin{equation}
    C_{\text{conv}}^{\text{dense}}
    = K_h\, \cdot K_w\, \cdot  \frac{C_{\text{in}}}{G}\, \cdot  C_{\text{out}}\,
      \cdot H_{\text{out}}\, \cdot W_{\text{out}}\,.
\end{equation}
The sparse weight MACs are $C_{\text{conv}}^{\text{sparse}} =
d_w^{(\ell)}\, C_{\text{conv}}^{\text{dense}}$. If a bias is present, its dense contribution is
$C_{\text{out}}\, H_{\text{out}}\, W_{\text{out}}$ additions, scaled by the
bias density $d_b^{(\ell)}$ for the sparse count.

\minisection{Linear layers.}
For a \texttt{Linear} layer with $F_{\text{in}}$ input features and
$F_{\text{out}}$ output features, the dense weight MACs are
\begin{equation}
    C_{\text{lin}}^{\text{dense}}
    = F_{\text{in}}\, \cdot F_{\text{out}}\,.
\end{equation}
Sparse weight MACs and bias additions follow the same density-scaling as for
convolutional layers.

\paragraph{Effective embedding dimension.}\label{app:effective_embedding}
Several acquisition functions (CoreSet, BADGE, SAAL, MaxHerding, UHerding)
operate on the penultimate-layer embedding of dimension~$e$. Under unstructured
weight pruning, some neurons in this layer may receive only zeroed inputs,
producing a constant activation across all data points. Such \emph{dead}
dimensions carry no discriminative signal yet still contribute to the cost of
pairwise distance computations if counted at face value. We therefore determine an
\emph{effective} embedding dimension $e_{\text{eff}} \leq e$ that reflects the
true embedding dimensionality of the representation after pruning.
To calculate $e_{\text{eff}}$ a batch of random Gaussian inputs is
passed through the model and the per-dimension variance of the resulting
embeddings is computed. A dimension is considered alive if its variance exceeds
zero (in practice a negligible numerical threshold suffices). The effective
dimension is then
\begin{equation}
    e_{\text{eff}} = \bigl|\{i : \operatorname{Var}_{x}[\phi(x)_i] > 0\}\bigr|
    \,,
\end{equation}
where $\phi(x) \in \mathbb{R}^{e}$ is the penultimate-layer representation.
Because the input is random noise, any dimension that remains constant is dead
by construction --- its output is determined entirely by bias terms unaffected by
the input --- rather than merely inactive on a particular data distribution.
Replacing $e$ with $e_{\text{eff}}$ in the acquisition cost formulas of
\Cref{tab:approx_acq_flops} yields tighter FLOPs estimates, especially at high
sparsity where a substantial fraction of embedding dimensions become dead.

\subsection{Inference \& DAL Score costs}\label{app:flops_scaling}

To coarsely bound the computational costs of acquisition functions $\mathcal{S}$ we introduce the following quantities, building
on the notation of \Cref{sec:method}. For readability we drop the iteration index
$(j)$ throughout this paragraph. Let $C_{\text{fwd}}(s)$ denote the FLOPs for a single
forward pass at sparsity~$s$; as a first-order approximation
$C_{\text{fwd}}(s) \approx (1 - s)\, C_{\text{fwd}}(0)$, i.\,e.\ linear savings
relative to the dense model. Let $e$ denote the dimensionality of the
penultimate-layer representation, $C$ the number of output classes, $\lvert D_U \rvert$ the
pool size, $\lvert D_L \rvert$ the labeled set size, and $B$ the per-round acquisition
budget. Every acquisition function begins with inference over $D_U$, contributing a
base cost of $\lvert D_U \rvert \cdot C_{\text{fwd}}(s)$. Beyond this shared term, methods that
rely on embeddings or gradient embeddings incur additional costs that scale with $e$,
$\lvert D_L \rvert$, and $B$. Most commonly, pairwise $\ell_2$ or inner-product distances introduce an $\mathcal{O}(e^2)$
term per candidate pair. Pruned models therefore improve acquisition efficiency along two
axes: first, through reduced forward-pass FLOPs at sparsity $s$, and second, through
a sparser penultimate-layer representation that lowers the \emph{effective}
embedding dimension $e_{\text{eff}} \leq e$ for all distance- or
similarity-based computations. We define $e_{\text{eff}}$ as the number of
embedding dimensions with nonzero variance across a batch of random inputs;
dimensions that are zeroed by pruning in the penultimate layer produce constant
outputs and contribute no discriminative signal.
\Cref{tab:approx_acq_flops} summarizes the leading-order cost approximation for each
acquisition function.

\subsection{Hyperparameter Settings \& Reproducibility}\label{app:hyperparams}
\Cref{tab:hyperparams} consolidates all training, pruning, and active learning
hyperparameters across the experimental scenarios evaluated in this work. Where a cell spans multiple columns, the value is shared across the corresponding
scenarios. All experiments were conducted on PyTorch~2.9 and torchvision version~0.24. 
Results can be reproduced from the configurations reported in \Cref{tab:hyperparams}. 
The complete source code, exact package versions, 
configuration files, and launch scripts are included in the supplementary material and will be released publicly upon acceptance.
Computations were distributed across a heterogeneous cluster of machines equipped with NVIDIA H100 and
RTX~3080\,Ti GPUs, with the larger backbones (ResNet-50, ConvNeXt\,v2, DeiT-S, DINO ViT-S) and the Places365
scenarios preferentially scheduled on the H100 nodes. In total, this work conducted more than $468$ DAL
experiments across 4 datasets, 5 architectures, 6 acquisition functions, 4 and training methods. 
Constituting a total of $\approx 6500$ individual model trainings.

\begin{table}[t]
    \centering
    \caption{Computational complexity approximation $\mathcal{O}(\cdot)$ for the acquisition phase of each
    method considered in this work. $C_{\text{fwd}}(s)$ denotes the forward-pass
    cost at sparsity $s$ and $e_{\text{eff}}$ is the effective penultimate-layer
    embedding dimension after pruning. The \emph{Subsampling} column indicates whether the method has been proposed with  random pool subsampling for tractability of larger pools.}
    \label{tab:approx_acq_flops}
    
    \begin{tabular}{ccc}
        \toprule
        \textbf{Acquisition Function $\mathcal{S}$} & \textbf{Subsampling} & \textbf{Computational Complexity $\mathcal{O}(\cdot)$}  \\
        \midrule
        Random & \xmark & $-$ \\
        \midrule
        Margin~SIGIR 1995 & \xmark & $\lvert D_U \rvert \cdot C_{\text{fwd}}(s)$ \\
        \midrule
        CoreSet~ICLR 2018 & \xmark & $(\lvert D_L \rvert + \lvert D_U \rvert) \cdot C_{\text{fwd}}(s) + B \cdot \lvert D_U \rvert \cdot e_{\text{eff}}^2$ \\
        \midrule
        BADGE~ICLR 2020 & \xmark & $\lvert D_U \rvert \cdot C_{\text{fwd}}(s) + B \cdot \lvert D_U \rvert \cdot e_{\text{eff}}^2$ \\
        \midrule
        SAAL~PMLR 2023 & \cmark & $5 \cdot \lvert D_U \rvert \cdot C_{\text{fwd}}(s) + B \cdot \lvert D_U \rvert \cdot e_{\text{eff}}^2$ \\
        \midrule
        MaxHerding~ECCV 2024 & \cmark & $(\lvert D_L \rvert + \lvert D_U \rvert) \cdot C_{\text{fwd}}(s) + (\lvert D_L \rvert + \lvert D_U \rvert)^2 \cdot e_{\text{eff}}^2$ \\
        \midrule
        UHerding~ICLR 2025 & \cmark & $(\lvert D_L \rvert + \lvert D_U \rvert) \cdot C_{\text{fwd}}(s) + (\lvert D_L \rvert + \lvert D_U \rvert)^2 \cdot e_{\text{eff}}^2$ \\
        \bottomrule
    \end{tabular}
\end{table}
\begin{table}[h!]
    \centering
    \caption{Full hyperparameter reference for reproducing all experiments. Each column
corresponds to one DAL scenario; rows are grouped by architecture,
pretraining, optimization, pruning, and active learning configuration.
LB = low budget, HB = high budget. KMeans + anchor modifies the standard initialization procedure by incorporating a set of fixed centers, e.g. anchors with 1 anchor for each class, the remaining elements from $\lvert D_L^{(0)}\rvert$ are chosen acording to the KMeans algorithm.}\label{tab:hyperparams}
    \resizebox{\linewidth}{!}{%
    \begin{tabular}{lcccccccc}
        \toprule
        & \multicolumn{2}{c}{\textbf{CIFAR-100}} & \multicolumn{2}{c}{\textbf{Imagewoof}}
        & \textbf{Tiny-ImageNet} & \multicolumn{2}{c}{\textbf{Places365}} \\
        \cmidrule(lr){2-3} \cmidrule(lr){4-5} \cmidrule(lr){6-6} \cmidrule(lr){7-8}
        & LB & HB & LB & HB & - & CNN & ViT \\
        \midrule
        \multicolumn{8}{l}{\textit{Architecture}} \\
        \quad Backbone
            & \multicolumn{2}{c}{ResNet-18} & \multicolumn{2}{c}{ResNet-18}
            & ResNet-18
            & \shortstack{ResNet-50 /\\ConvNext v2}
            & \shortstack{DeiT-S /\\DINO ViT-S} \\
        \quad Input resolution 
            & \multicolumn{2}{c}{$32{\times}32$} & \multicolumn{2}{c}{$224{\times}224$}
            & $64{\times}64$
            & \multicolumn{2}{c}{$224{\times}224$} \\
        \midrule
        \multicolumn{8}{l}{\textit{Pretraining}} \\
        \quad Method
            & \multicolumn{2}{c}{SimCLR} & \multicolumn{2}{c}{SimCLR}
            & SimCLR
            & \multicolumn{2}{c}{ImageNet} \\
        \quad Paradigm
            & \multicolumn{2}{c}{Self-sup.} & \multicolumn{2}{c}{Self-sup.}
            & Self-sup.
            & \shortstack{Sup. / \\Self-sup.}
            & \shortstack{Sup. / \\Self-sup.} \\
        \midrule
        \multicolumn{8}{l}{\textit{Optimization}} \\
        \quad Optimizer
            & \multicolumn{2}{c}{AdamW} & \multicolumn{2}{c}{AdamW} & AdamW & \multicolumn{2}{c}{AdamW} \\
        \quad Weight decay
            & \multicolumn{2}{c}{$10^{-4}$} & \multicolumn{2}{c}{$10^{-4}$} & $10^{-4}$ & \multicolumn{2}{c}{$10^{-4}$} \\
        \quad Batch size
            & $128$ & $32$ & $64$ & $32$
            & $64$ & \multicolumn{2}{c}{$128$} \\
        \quad Max epochs (1st stage)
            & \multicolumn{2}{c}{$150$} & \multicolumn{2}{c}{$300$}
            & $150$ & \multicolumn{2}{c}{$45$} \\
        \quad Max epochs (2nd stage)
            & \multicolumn{2}{c}{$30$} & \multicolumn{2}{c}{$60$}
            & $30$ & \multicolumn{2}{c}{$15$} \\
        \quad Head warmup
            & \multicolumn{2}{c}{-} & \multicolumn{2}{c}{-}
            & - & \multicolumn{2}{c}{$3$} \\
        \quad LR warmup epochs
            & \multicolumn{2}{c}{$10$} & \multicolumn{2}{c}{$20$}
            & $10$ & \multicolumn{2}{c}{$5$} \\
        \quad LR warmup range
            & \multicolumn{2}{c}{$10^{-6} \to 10^{-3}$}
            & \multicolumn{2}{c}{$10^{-6} \to 10^{-3}$}
            & $10^{-6} \to 10^{-3}$
            & \multicolumn{2}{c}{$10^{-6} \to 10^{-3}$} \\
        \quad LR schedule
            & \multicolumn{2}{c}{Cosine}
            & \multicolumn{2}{c}{Cosine}
            & Cosine
            & \multicolumn{2}{c}{Cosine} \\
        \quad LR schedule range
            & \multicolumn{2}{c}{$10^{-3} \to 5 \cdot 10^{-5}$}
            & \multicolumn{2}{c}{$10^{-3} \to 5 \cdot 10^{-5}$}
            & $10^{-3} \to 5 \cdot 10^{-5}$
            & \multicolumn{2}{c}{$10^{-3} \to 5 \cdot 10^{-5}$}\\
        \quad Convergence threshold
            & \multicolumn{2}{c}{$99\%$} & \multicolumn{2}{c}{$97.5\%$}
            & $99\%$ & \multicolumn{2}{c}{$99\%$} \\
        \midrule
        \multicolumn{8}{l}{\textit{Pruning}} \\
        \quad Min. active parameter fraction $1 - s$
            & \multicolumn{2}{c}{5\%} & \multicolumn{2}{c}{5\%}
            & 5\% & 30\% & 40\% \\
        \quad Pruning rate $p$
            & \multicolumn{2}{c}{0.66} & \multicolumn{2}{c}{0.66}
            & 0.66 & 0.25 & 0.15 \\
        \quad Reactivation fraction
            & \multicolumn{2}{c}{0.1} & \multicolumn{2}{c}{0.1}
            & 0.1 & \multicolumn{2}{c}{0.05} \\
        \quad Rewind epoch $r$
            & \multicolumn{2}{c}{5} & \multicolumn{2}{c}{10}
            & 5 & \multicolumn{2}{c}{ - } \\
        \midrule
        \multicolumn{8}{l}{\textit{Active Learning}} \\
        \quad Initial Selection 
            & \multicolumn{2}{c}{KMeans + anchor} & \multicolumn{2}{c}{KMeans + anchor} 
            & Stratified Random & \multicolumn{2}{c}{Stratified Random} \\
        \quad Initial pool $|D_L^{(0)}|$
            & $200$ & $2000$ & $100$ & $500$
            & $1000$ & \multicolumn{2}{c}{$5475$} \\
        \quad Budget per round $B$
            & $100$ & $2000$ & $50$ & $500$
            & $200$ & \multicolumn{2}{c}{$3650$} \\
        \quad DAL iterations $A$
            & \multicolumn{2}{c}{$9$} & \multicolumn{2}{c}{$9$}
            & $20$ & \multicolumn{2}{c}{$9$} \\
        \quad Acquisition functions
            & \multicolumn{4}{c}{\shortstack{Margin, CoreSet, BADGE,\\SAAL, MaxHerding, UHerding}}
            & \shortstack{Margin, CoreSet, \\ BADGE}
            & \multicolumn{2}{c}{Margin} \\
        \quad Seeds
            & \multicolumn{2}{c}{$10$} & \multicolumn{2}{c}{$10$} & 15 & \multicolumn{2}{c}{$3$} \\
        \bottomrule
    \end{tabular}}%
\end{table}

UHerding~\citep{bae2025uncertainty} originally calibrates its uncertainty
estimates via temperature scaling by a dual model training stage. First a model is trained with a held-out test set 
split from the available labeled set $D_L^{(j)}$ to estimate the temperature parameter $\tau$. Afterwards the
model is retrained again on the complete labeled set and $\tau$ is applied without further calibration.
To maintain a uniform computational baseline across all acquisition functions, we use the default recommended
temperature $\tau=1$ and omit the additional calibration training, ensuring that
no acquisition function incurs extra model training beyond the shared DAL loop.

\begin{figure}[b]
  \centering
  \includegraphics[width=.9\linewidth]{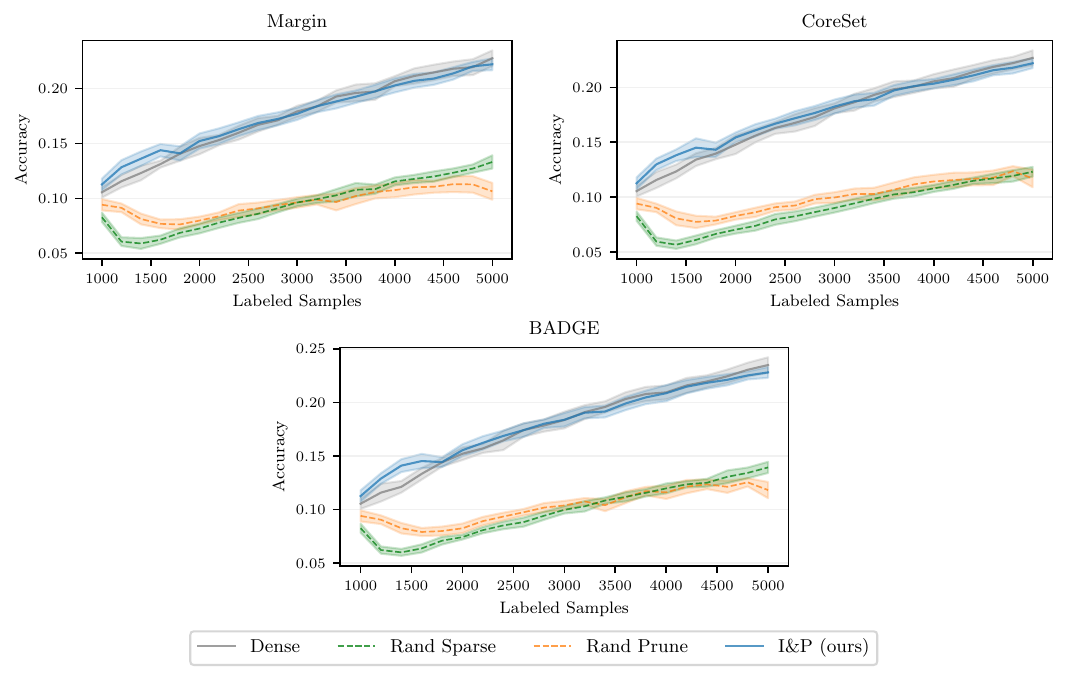}
  \caption{Mean Accuracy on Tiny-ImageNet for Margin, CoreSet, and
      BADGE acquisition function $\mathcal{S}$ over \num{20} DAL iterations with $\lvert D_L^{(0)} \rvert=1000$ stratified samples and an annotation budget of \num{200} samples per iteration.
      Shaded regions indicate $\pm 1$ std over 15 seeds.}
  \label{fig:curves_tinyimagenet}
\end{figure}

\subsection{Per-Iteration Learning Curves}\label{app:learning_curves}

We report mean test accuracy on Tiny-ImageNet over the DAL iterations with shaded regions indicating $\pm 1$ std over seeds for all four training strategies (Dense, Rand Sparse, Rand Prune, \method).
We can observe a consistent pattern in \Cref{fig:curves_tinyimagenet} that also generalizes to CIFAR-100, Imagewoof. In the low-budget regime, and during the early iterations of the high-budget regime, \method matches or slightly outperforms the Dense baseline, which we attribute to the slightly lower sparsity levels ($< 80 \%$) in the beginning combined with increased regularization effects from the sparse models. 
In the later iterations of the high-budget regime, \method falls marginally below Dense, typically by less than $\epsilon < 0.5\%$ absolute accuracy --- a gap we regard as practically negligible given the sparsity level reached at that point. In contrast, both random baselines (Rand Sparse, Rand Prune) trail Dense and \method by a substantial margin throughout, confirming a trivial random sparsity is insufficient for this performance level. These trends are remarkably stable and translate across acquisition functions. Suggesting that the compatibility of \method with the DAL loop is largely orthogonal to the choice of acquisition function.

\begin{figure}
  \centering
  \includegraphics[width=.9\linewidth]{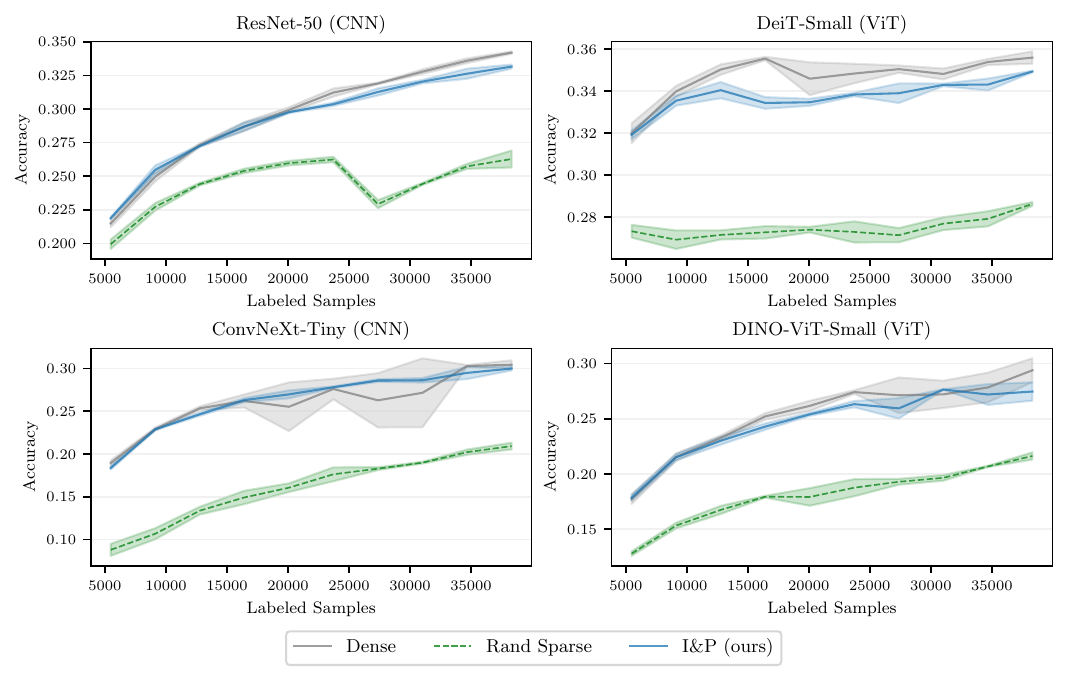}
  \caption{Mean Accuracy on Places365 (active finetuning) for Margin acquisition function $\mathcal{S}$ over \num{9} DAL iterations with $\lvert D_L^{(0)} \rvert=5475$ class-balanced samples and an annotation budget of \num{3650} samples per iteration. Rows correspond to different ImageNet-pretrained architectures spanning CNN and Transformer families with supervised and self-supervised pretraining.
      Shaded regions indicate $\pm 1$ std over 3 seeds.}
  \label{fig:curves_places365}
\end{figure}

DeiT-Small visualized in \Cref{fig:curves_places365} is the only architecture and setting that deviates significantly. 
In our evaluation (3 seeds) the test accuracy
temporarily decreases during the DAL cycle before recovering at later
iterations. Notably, this dip affects both the
Dense baseline and \method, indicating that it is not entirely an artifact of pruning.
We hypothesize that the mid-cycle labeled set occupies an unfavorable regime
for domain transfer: it contains enough samples to overfit to Places365 and
erode the pretrained ImageNet features, but not yet enough to generalize well
to the new domain. As more data is acquired in later iterations, generalization
recovers and accuracy resumes its upward trend.

\subsection{Accuracy vs. Cumulative Training MACs}\label{app:accuracy_vs_flops}

\begin{figure}
  \centering
  \includegraphics[width=0.9\linewidth]{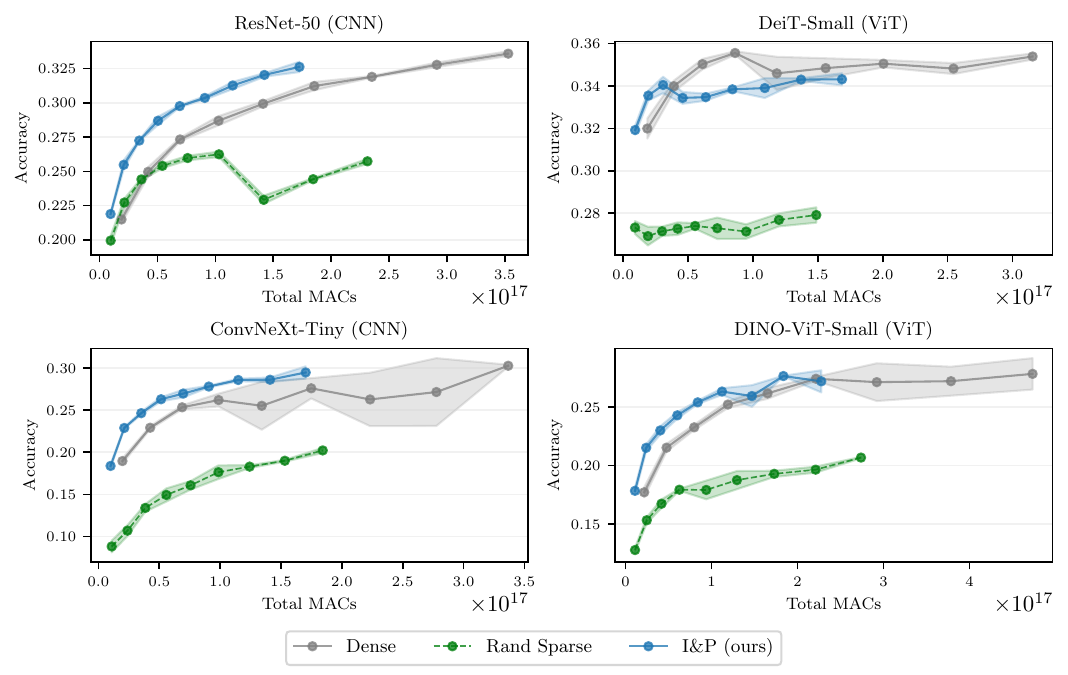}
  \caption{Test accuracy vs.\ cumulative training MACs on Places365
      (active finetuning) with Margin acquisition. Rows correspond to
      different ImageNet-pretrained architectures spanning CNN and
      Transformer families with supervised and self-supervised
      pretraining.}
  \label{fig:flops_places365}
\end{figure}

\begin{figure}
  \centering
  \includegraphics[width=\linewidth]{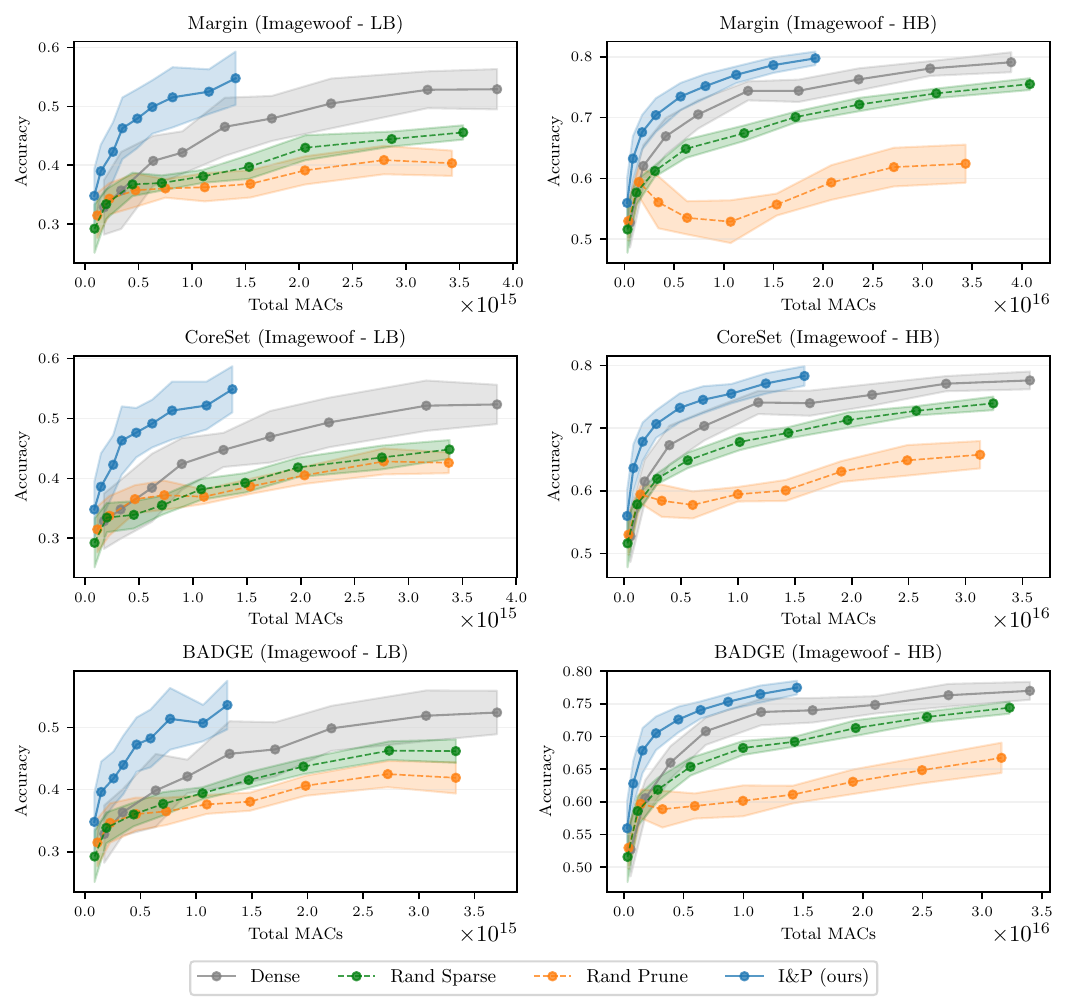}
  \caption{Test accuracy vs.\ cumulative training MACs on Imagewoof for
      Margin, CoreSet, and BADGE. Left column: low budget (LB); right
      column: high budget (HB). Each marker corresponds to one DAL iteration.}
  \label{fig:flops_imagewoof_part1}
\end{figure}

\begin{figure}
  \centering
  \includegraphics[width=\linewidth]{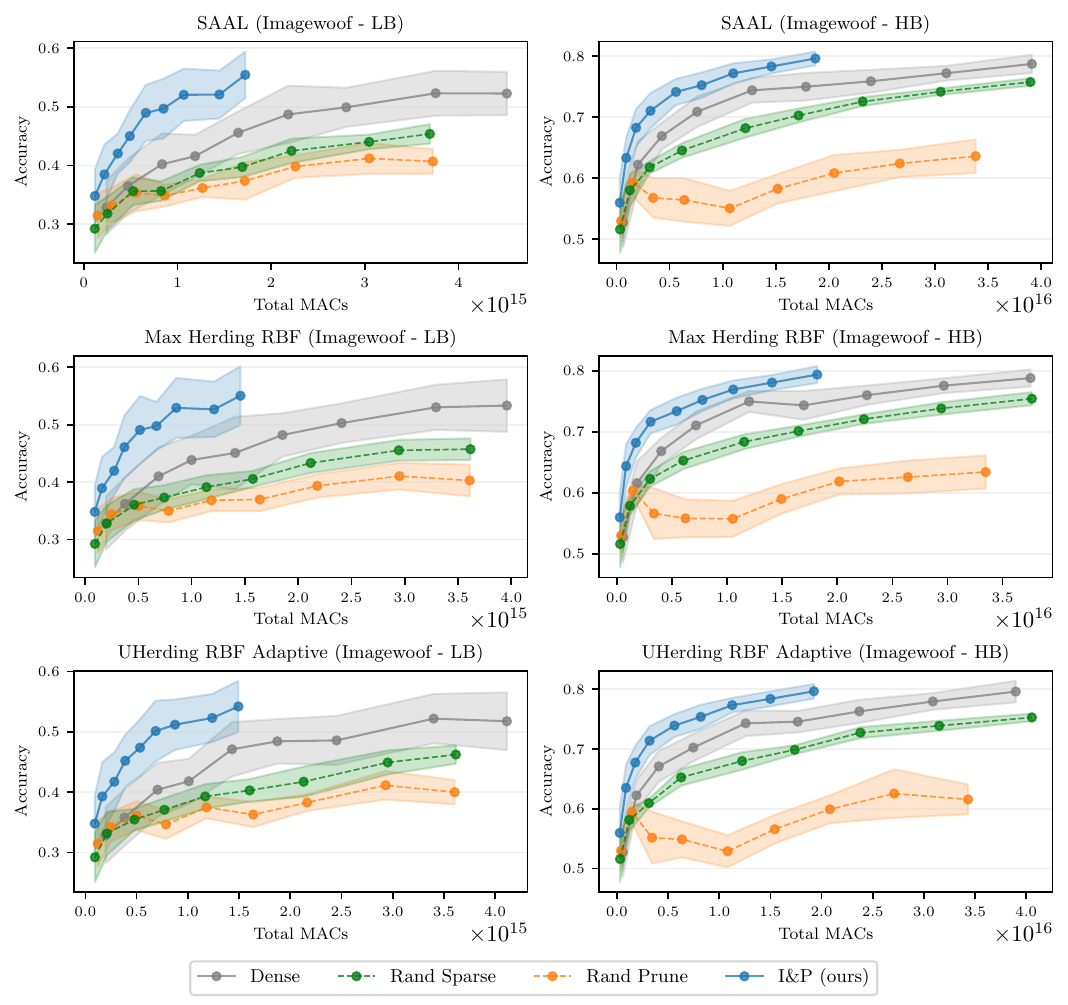}
  \caption{Test accuracy vs.\ cumulative training MACs on Imagewoof for
      SAAL, MaxHerding, and UHerding. Left column: low budget (LB); right
      column: high budget (HB). Each marker corresponds to one DAL iteration.}
  \label{fig:flops_imagewoof_part2}
\end{figure}

\Cref{fig:flops_imagewoof_part1} and \Cref{fig:flops_imagewoof_part2} depict mean accuracy with one standard deviation as a
function of cumulative MACs, aggregating training, finetuning, and acquisition function costs across the full DAL cycle.
The trends are consistent across all acquisition functions and
dataset settings: \method achieves comparable accuracy to Dense while saving
between $40$ and $60\%$ of the total computation. Both random baselines fall
well below Dense in accuracy and, due to their exhausted patience budgets and
prolonged training times, sometimes even exceed Dense in cumulative MACs. While
the qualitative trend is stable, the nominal scale of MACs differs substantially
between scenarios (\Cref{fig:flops_places365}) and budget regimes, reflecting the varying input resolutions,
model sizes, and label pool growth rates.

\end{document}